\documentclass{article}

\usepackage{times}
\usepackage{soul}
\usepackage{url}
\usepackage[small]{caption}
\usepackage{amsmath}
\usepackage{amsthm}
\usepackage[switch]{lineno}
\usepackage{longtable}
\usepackage{supertabular}
\usepackage{multirow}
\usepackage{xcolor}
\usepackage{hyperref}
\usepackage{enumitem}
\setlist{leftmargin=1cm}

\usepackage[in]{fullpage}  
\usepackage{arxiv} 
\usepackage{mathpazo} 

\usepackage[utf8]{inputenc} 
\usepackage[T1]{fontenc}    
\usepackage{url}            
\usepackage{booktabs}       
\usepackage{amsfonts}       
\usepackage{microtype}      
\usepackage{lipsum}         

\usepackage{graphicx}  
\usepackage{float} 
\usepackage{subfigure} 
\usepackage{algorithm}
\usepackage{algorithmic}
\usepackage{tabulary}

\usepackage{amssymb}
\usepackage{pifont}

\usepackage{minitoc}

\newcommand{\methodname}{KALM}

\def\gA{{\mathcal{A}}}

\def\gD{{\mathcal{D}}}

\def\gG{{\mathcal{G}}}

\def\gM{{\mathcal{M}}}

\def\gP{{\mathcal{P}}}

\def\gR{{\mathcal{R}}}
\def\gS{{\mathcal{S}}}

\hypersetup{
    colorlinks=true,
    citecolor=blue,
    linkcolor=blue,
    citecolor=orange
}

\title{Knowledgeable Agents by Offline Reinforcement Learning from Large Language Model Rollouts}

\usepackage{authblk}
\author{%
  Jing-Cheng Pang\textsuperscript{\rm 1,2}, 
  Si-Hang Yang \textsuperscript{\rm 1,2}, 
  Kaiyuan Li\textsuperscript{\rm 1,2}, 
  Jiaji Zhang\textsuperscript{\rm 1}, 
  Xiong-Hui Chen\textsuperscript{\rm 1,2}, 
  
  Nan Tang\textsuperscript{\rm 1,2}, 
  Yang Yu\textsuperscript{\rm 1,2,$\diamond$}\\
  \textsuperscript{\rm 1} National Key Laboratory for Novel Software Technology, Nanjing University, China \\ \& School of Artificial Intelligence, Nanjing University, China \\
  \textsuperscript{\rm 2}Polixir.ai \\
  \textsuperscript{$\diamond$} Corresponding: yuy@nju.edu.cn
}

\date{}
\begin{document}

\doparttoc %
\faketableofcontents %

\maketitle

\begin{abstract}
  Reinforcement learning (RL) trains agents to accomplish complex tasks through environmental interaction data, but its capacity is also limited by the scope of the available data. To obtain a knowledgeable agent, a promising approach is to leverage the knowledge from large language models (LLMs). Despite previous studies combining LLMs with RL, seamless integration of the two components remains challenging due to their semantic gap. This paper introduces a novel method, Knowledgeable Agents from Language Model Rollouts (\methodname), which extracts knowledge from LLMs in the form of imaginary rollouts that can be easily learned by the agent through offline reinforcement learning methods. The primary challenge of \methodname~lies in LLM grounding, as LLMs are inherently limited to textual data, whereas environmental data often comprise numerical vectors unseen to LLMs. To address this, \methodname~fine-tunes the LLM to perform various tasks based on environmental data, including bidirectional translation between natural language descriptions of skills and their corresponding rollout data. This grounding process enhances the LLM's comprehension of environmental dynamics, enabling it to generate diverse and meaningful imaginary rollouts that reflect novel skills. Initial empirical evaluations on the CLEVR-Robot environment demonstrate that \methodname~enables agents to complete complex rephrasings of task goals and extend their capabilities to novel tasks requiring unprecedented optimal behaviors. \methodname~achieves a success rate of 46\% in executing tasks with unseen goals, substantially surpassing the 26\% success rate achieved by baseline methods. Furthermore, \methodname~effectively enables the LLM to comprehend environmental dynamics, resulting in the generation of meaningful imaginary rollouts that reflect novel skills and demonstrate the seamless integration of large language models and reinforcement learning.
\end{abstract}

\section{Introduction}

Acquiring new skills autonomously is a hallmark of the intelligence. Reinforcement learning (RL) has emerged as a powerful mechanism for training intelligent agents to perform complex tasks in interactive environments \cite{alphago,alphastar,dqn}. This approach enables an agent to learn by collecting interacting data from the environments, or a static dataset of environmental interactions without real-time interactions. 
While RL has shown promise in many challenging tasks, its efficacy is often questioned due to the limitations of the learned policy, which is inherently constrained by the scope of the skills presented within the data. Consequently, policies may not generalize well to new situations, leading to bad performance in unencountered scenarios. For instance, a policy trained to \emph{move an object to the left} may fail when required to \emph{move it to the right}, despite the they are highly similar tasks.

On the other hand, recent advancements in large language models (LLMs) have opened up new opportunities for intelligent agents to acquire knowledge and solve general textual problems, including dialogue, reasoning, and mathematical problem-solving \cite{dialog1,lamai,vlm1,cot,RLC}. 
LLMs, trained on extensive text corpora, encapsulate a broad spectrum of world knowledge. To leverage such knowledge, researchers have employed pre-trained LLMs to construct knowledge bases \cite{llm_as_knowledge_base}, decompose intricate tasks \cite{saycan}, and augment programming efforts \cite{assist_program}. Furthermore, evidence suggests that LLMs can extend beyond text-based tasks, aiding in decision-making within interactive, embodied environments \cite{extract_action_know,llm_gen_policy,talar}. This suggests that LLMs may offer a form of general knowledge that could enhance offline RL by addressing the constraints of dataset-dependent policies.

\begin{figure}[t]
    \centering
    \includegraphics[width=1\linewidth]{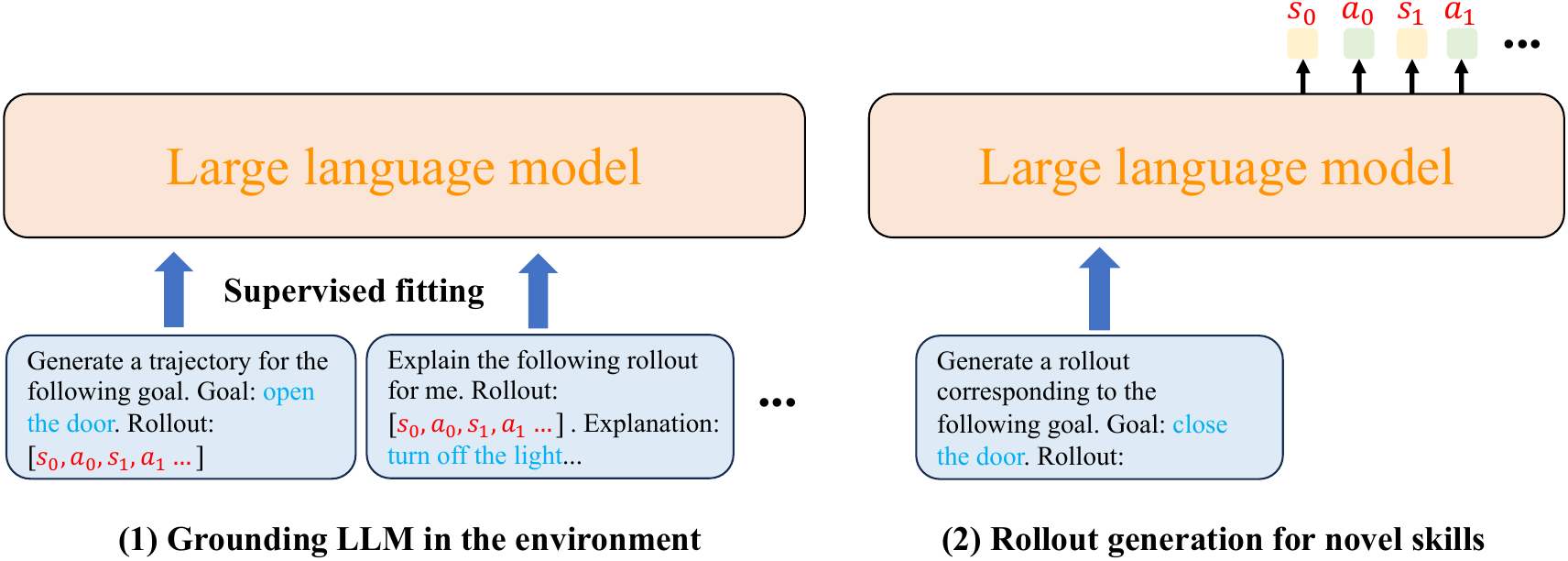}
    \caption{Illustration of \methodname~utilizing LLM to generate environmental rollouts. \textbf{(1)} Grounding phase that fine-tunes LLM with supervised fitting on the environmental data. \textbf{(2)} Generation phase that prompts LLM to generate data for novel skills. \methodname~modifies the input/output layer of LLM, enabling it to process and interpret non-textual data.}
    \label{fig:world_model_illustration}
\end{figure}

This study investigates the development of a knowledgeable policy capable of adapting to novel situations through the integration of LLMs. We introduce \underline{K}nowledgeable \underline{A}gent from \underline{L}anguage \underline{M}odel rollouts (\methodname) method, which employs a pre-trained LLM to create synthetic rollouts that simulate the performance of new skills. Figure \ref{fig:world_model_illustration} depicts the process by which \methodname~utilizes an LLM to generate these environmental rollouts. 
The motivation behind \methodname~is that LLMs, with their extensive repository of knowledge, are ideally suited for generating rollouts that encompass a variety of skills. However, a key challenge is that LLMs are inherently designed to process textual data. To address this, \methodname~applies two techniques to ground LLM in the environment: adapting the LLM's architecture to interpret environmental observations and actions, and fine-tuning the LLM with environmental data to predict execution rollouts and translate between natural language goals and their corresponding rollouts. \methodname~then uses the LLM to generate imaginary rollouts for a range of skills, including those not present in the original dataset, and integrates these with offline RL to enhance skill acquisition. This novel combination of LLM-generated imaginary rollouts with offline RL marks potentially yields more versatile and informed agents.

Our contributions are summarized as follow: We emonstrate the seamless integration of large
language models and reinforcement learning, enriching the RL training with a broader skill set. We introduce the \methodname~method, which implements this concept by grounding LLMs within the environment to generate rollouts of various skills, thereby facilitating skill acquisition via offline RL. We also develop an effective technique for aligning the LLM with multimodal data inputs, such as text and vectors. Finally, we provide empirical evidence of \methodname's efficacy. Our experiments with the CLEVR-Robot task indicate that \methodname~can fine-tune the LLM to generate effective rollouts and acquire new skills, achieving a 46\% success rate on tasks described in novel natural language goals, significantly outperforming the baseline offline RL method's 26\% success rate.

\section{Related Work}

\subsection{Offline Reinforcement Learning}

Offline RL \cite{offlinerlsurvey,bcq} enables agents to learn from a static dataset of pre-collected experiences, rather than requiring real-time interaction with the environment. The core challenge in offline RL is to derive effective policies from a dataset that may be biased or has limited data due to the behavior policies that generate it. Advancements in this domain have introduced novel techniques such as importance sampling, conservative policy evaluation, and representation learning to address these challenges \cite{cql,uwac,iql,HIDIL}. Despite its successes, offline RL is inherently limited by the quality and diversity of the dataset. If the dataset lacks certain experiences or skills, the agent may fail to perform adequately in those scenarios. Approaches like MAPLE \cite{maple} and ReDM \cite{policy_rehaersal} attempt to overcome this by training diverse world models to simulate a range of data, thereby enhancing policy robustness in unfamiliar scenarios. These models, however, are typically learned from scratch and prioritize extensive data coverage, which may not align with real-world data distributions. In contrast, our paper leverages a pre-trained LLM to facilitate the generation of higher-quality imaginary rollouts, potentially addressing the limitations of previous methods.

\subsection{Large Language Models}

Large language models (LLMs) exhibit remarkable proficiency in processing and comprehending natural language \cite{gpt4,llama2,chatglm}. More exciting, their capabilities extend to tasks beyond basic language understanding problems, including dialogue \cite{dialog1,dialog2,lamai}, multimodal vision-language tasks \cite{vlm1}, logical reasoning \cite{cot,RLC}, and mathematical problem-solving \cite{math}. These models are characterized by their deep neural networks, which consist of hundreds of layers and billions of parameters, enabling them to capture the nuances of natural language with unprecedented depth \cite{transformer,bert}. The pre-training process involves exposing the models to vast corpora of text, allowing them to learn a wide range of linguistic patterns and commonalities before being fine-tuned for specific tasks. Recent advancements, such as GPT-4 \cite{gpt4}, have pushed the boundaries further, not only in terms of scale but also through more sophisticated training techniques and broader contextual understanding \cite{gpt2}. While their primary utility has been within the realm of text, there is an emerging trend to explore the capabilities of LLMs in interactive environments \cite{ground_llm_in_env,llm_gen_policy,llm_planner}, capitalizing on their vast world knowledge. Besides, this work investigates the utility of LLMs to generate environmental data.

\subsection{Integrating LLMs and RL}

How to effectively leverage LLMs for decision-making in interactive tasks is an important area of study. The central strategy is to capitalize on the extensive prior knowledge embedded within LLMs. Research in this domain has taken several approaches. One approach involves decomposing the complex tasks, and utilize LLMs to generate high-level plans, which are then executed by a low-level controller \cite{saycan,innermono,llm_planner}. Another approach uses LLMs to design the reward function \cite{reward_design1,reward_design2,reward_design3}, which streamlines the otherwise laborious process of reward function formulation. Additionally, there is a trend towards the direct application of LLMs as decision-making agents, integrating them with RL to facilitate direct interaction with various environments. Examples include GALM \cite{ground_llm_in_env} and TWOSOME \cite{twosome}, which ground LLMs in text-based games, and LLaRP \cite{llm_gen_policy}, which adapts a conventional LLM for physical tasks and develops an RL-trained policy that enables direct action output from LLMs. Different from these studies, in this work, we propose a novel solution that utilizes LLM to generate imaginary rollouts for novel skill learning.

\section{Preliminary}

\paragraph{Reinforcement learning.} 

In this paper, we consider a RL task that agent needs to complete tasks assigned by natural language. We model the environment as a goal-augmented Markov Decision Process (MDP) \cite{puterman2014markov, sutton2011reinforcement}, represented by the tuple $\gM = \left( \gS, \gA, \gP, \gR, \gamma, \gG \right)$. In this tuple, $\gS$ denotes the state space, $\gA$ the action space, $\gP$ the transition function of the environment, $\gR$ the reward function that evaluates the quality of the agent's action, $\gamma$ the discount factor which balances the immediate and future rewards, and $\gG$ the set of natural language goals. 
A policy $\pi: \gS \times \gG \rightarrow \Delta (\gA)$ defines the strategy the agent employs, mapping states and goals to a distribution over possible actions.
The interaction between the RL agent and the environment proceeds as follows: at each timestep $t$, the agent observes a state $s_t$ and an goal $G$ from the environment. It then selects an action $a_t$ based on the policy $\pi (\cdot |s_t, G)$. Upon executing this action, the agent receives a reward $r(s_t, a_t, G)$ and the environment transitions to a new state $s_{t+1}$ according to the transition function $\gP(\cdot |s_t, a_t)$. The objective of RL is to find a policy that maximizes the expected sum of rewards over time. In this study, we call the state, action data of the environment as the \emph{environmental data}.

\paragraph{Large language model.}

We employ a large language model $\gM$, an autoregressive text generation model that predicts future tokens in a sequence. 
An LLM predicts the next token $l_{t+1}=\gM(E_T(l_0),\cdots,E_T(l_t))$ conditioned on all prior sequence of tokens, where $E_T$ is the token embedding layer that converts token into a D-dimensional embedding, $l_t \in \Sigma$ and $\Sigma$ denotes the vocabulary of the LLM. 
In this study, the LLM needs to understand not only textual data, but the state and action data of an environment; these data formats differs from language tokens. Specifically, the LLM's input layer converts text tokens into vector embeddings $e_k=E_T(l_k)\in \mathbb{R}^D$, and its output layer classifies tokens. To facilitate the processing of state and action data, we have modified the LLM's architecture by replacing its original input and output layers with multi-layer perceptrons (MLPs), thereby enabling the integration of non-textual environmental data.

\section{Method}

\label{sec:method}

\begin{figure}[t]
    \centering
    \includegraphics[width=1\linewidth]{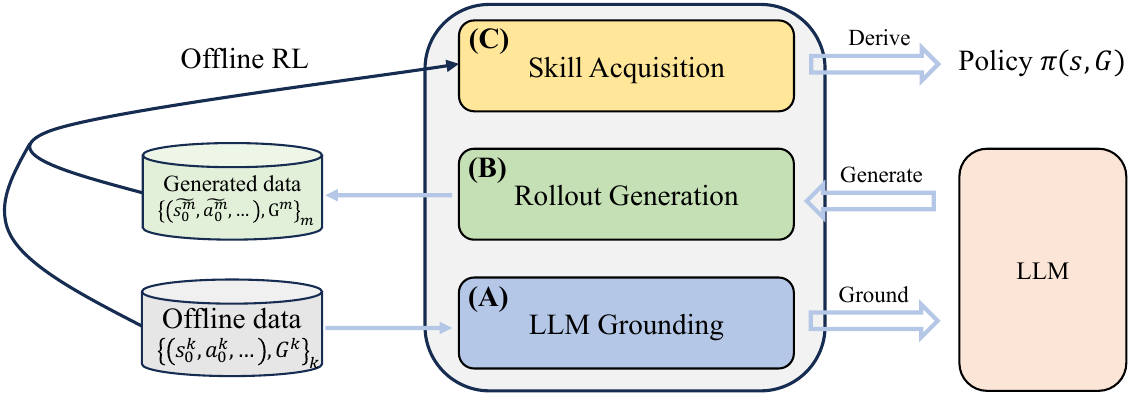}
    \caption{Overall procedure of \methodname, consisting of three key modules: (A) LLM grounding module that grounds LLM in the environment and aligns LLM with inputs of environmental data, (B) Rollout generation module that prompts the LLM to generate data for novel skills and (C) Skill Acquisition module that trains the policy with offline RL. Finally, \methodname~derives a policy that trained on both offline data and imaginary data.}
    \label{fig:overall_framework}
\end{figure}

This section presents the proposed Skill Acquisition with Language Model (\methodname) method for intelligent agents to acquire novel skills. Figure \ref{fig:overall_framework} shows the overall running procedure of \methodname, which comprises three important components: (1) LLM grounding module that enables LLM to understand the elements of the environment, (2) rollout generation module that generates imaginary rollouts for novel skills, and (3) Skill acquisition module that trains the policy with offline RL. We first give a formal definition of the problem.

\subsection{Problem Formulation}
Consider we have an offline dataset $\gD$ composed of paired goals and rollouts, symbolized as $\{G^k, (s_0^k,a_0^k,s_1^k,a_1^k,\cdots) \}_{k=1}^K$. In this context, $(s_0^k,a_0^k,s_1^k,a_1^k,\cdots)$ is a rollout for executing certain task in the environment, detailing the sequence of states and actions $(s_i^k,a_i^k)$ required to complete the goal $G^k$. Typically, $G^k$ can reflect different skills for intelligent agent. The primary objective here is to obtain a policy that achieve high rewards on unseen goal distributions, represented as $\gG'$, thus ensuring its ability beyond the constraints of the offline dataset.

\subsection{LLM Grounding with Instruction-following Fine-tuning}
The first step of \methodname~is to ground LLM in the environment. This process is designed to equip the LLM with the capability to interpret the meaning of states, actions, dynamics and rollout data within the given environments. To implement this idea, we fine-tune the LLM using the offline dataset $\gD$ to perform four different tasks in supervised manner:

\begin{itemize}[leftmargin=1cm]
    \item Dynamics prediction: The LLM predicts environmental dynamics changes. Given the current state $s_t$, and action $a_t$, the LLM forecasts the subsequent state.
    \item Rollout explanation: The LLM explains a given rollout sequence $s_0, a_0, s_1,\cdots$ using natural language.
    \item Rollout generation: The LLM generates a rollout sequence that aligns with a specified goal $G$.
    \item Consequence prediction: The LLM predicts the terminal state resulting from completing a goal $G$ starting from an initial state $s_0$.
\end{itemize}

Here, \methodname~models the LLM grounding problem as a instruction-following problem: LLM demonstrates great performance in following given natural language instruction to generate the desired answer. In this way, we can adjust the instruction prompt input to the LLM to better utilize the LLM and specify the generation objective for LLM. The prompts we use are presented in Appendix \ref{appsec:prompt}.

\begin{figure}[t]
    \centering
    \includegraphics[width=0.95\linewidth]{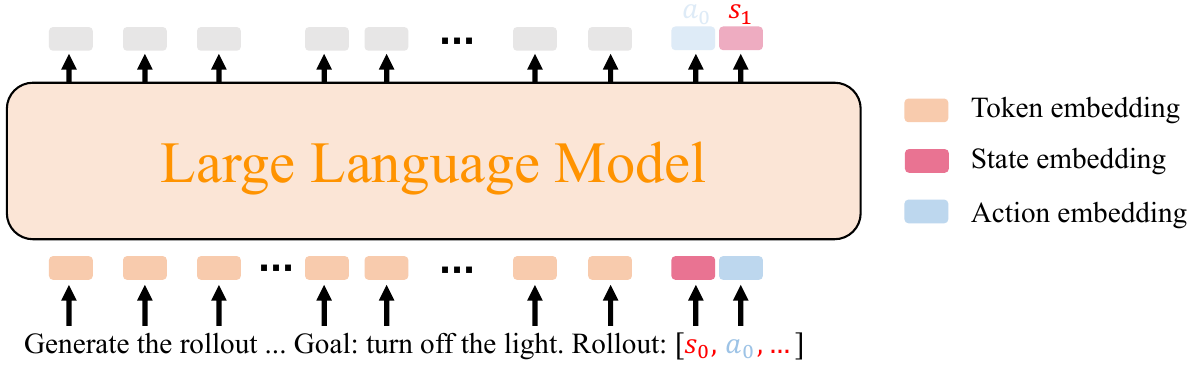}
    \caption{\methodname~utilizes a pre-trained LLM as the backbone model, but adapts the architecture the LLM to process non-textual data. It accepts a variety of data types, including \textcolor{red}{state}, \textcolor{cyan}{action}, and \textcolor{orange}{text} token. \methodname~employs distinct embedding layers to convert these inputs into same-dimension embeddings. For outputs, the LLM uses different output heads to output the predicted state, action and text.}
    \label{fig:network_arch}
\end{figure}

Given that LLMs are inherently trained on textual data to process and predict sequences of tokens, they can not be directly utilized to generate imaginary rollouts. To overcome this limitation, \methodname~introduces modifications to the LLM's network architecture. As shown in Figure \ref{fig:network_arch}, we use a pre-train LLM as the backbone model and enhance it with additional layers to handle environmental data. For inputs such as states and actions, we incorporate learnable embedding modules, $E_S: \mathcal{S} \rightarrow \mathbb{R}^D$ and $E_A: \mathcal{A} \rightarrow \mathbb{R}^D$, which transform these inputs into embeddings of the same dimensionality as the token embeddings. For outputs, we employ learnable modules, $O_S: \mathbb{R}^D \rightarrow \mathcal{S}$ and $O_A: \mathbb{R}^D \rightarrow \mathcal{A}$, which map the LLM's output into state space $\mathcal{S}$ or action space $\mathcal{A}$. This framework can be easily extended to tasks involving visual observations by integrating appropriate neural network architectures.

\subsection{Rollout Generation with Novel Skill Prompt}

After fine-tuning the LLM with environmental data, it acquires the capability to interpret the states, actions, and dynamics within an environment. Given that LLMs possess a broad spectrum of world knowledge, it is potential that they can generate imaginary rollout for a diverse range of new skills. To this end, we employ the optimized model to generate imaginary rollouts with specific prompts, such as "Generate a rollout for the following goal: [GOAL]," where "[GOAL]" is a placeholder for various target objectives that reflect diverse skills. Following prior research that focuses on policy generalization \cite{llm_gen_policy}, as documented in the literature, we measure the novelty of skills along two dimensions:

\begin{itemize}
    \item Paraphrastic Robustness: This dimension assesses the model's consistency in optimal behavior when faced with linguistically diverse goals that share the same underlying intent as previously encountered goals. It includes alternative phrasings for identical actions or re-expressing the name of the objects.
    \item Novel Task Generalization: Here, we investigate the model's proficiency in performing tasks that demand the formulation of new optimal behaviors. Such tasks or skills never occur in the offline dataset. For instance, if the offline dataset includes tasks related to making a robot walk, a novel task might involve enabling the robot to run. 
    These tasks necessitate a comprehensive understanding by the LLM of the objects and the environmental dynamics to generate effective state-action sequences.
\end{itemize}

We elaborate how we construct novel tasks that align with these dimensions in Section \ref{sec:exp_setting}. For illustrative examples of novel goals, please refer to the Appendix \ref{appsec:exm_novel_goal}.

\subsection{Skill Acquisition via Offline Reinforcement Learning}

\methodname~employs offline RL approach to train a policy $\pi(\cdot|s,G)$, utilizing both the existing offline dataset and imaginary rollouts generated from LLM, with a same proportion of two sources of rollouts. To build policy network, BERT \cite{bert} serves as the encoder for processing natural language goals, due to its proficiency in text encoding. The encoded goals are integrated with the state representations to form the input for the policy network.
For policy optimization, \methodname~is compatible with any offline RL algorithm, such as behavior cloning and CQL \cite{cql}, leveraging the combined data from the offline dataset and generated rollouts. It is convenient for the incorporation of continuously generated imaginary rollouts into the training process, making the proposed framework more flexible. Furthermore, we conduct a comprehensive analysis of \methodname's performance when integrated with a variety of offline RL algorithms, the results of which are detailed in Section \ref{sec:exp_analysis}.

\begin{figure}[t]
    \centering
    \subfigure[Beginning of a task]{\includegraphics[width=0.22\linewidth]{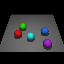}}
    \subfigure[Termination of a task]{\includegraphics[width=0.22\linewidth]{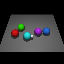}}
    \caption{A visualization of CLEVR-Robot environment in our experiments. The agent (silverpoint) manipulates five movable balls to reach a specific configuration. An example of natural language goal in the offline dataset is: \emph{Can you move the \textcolor{violet}{purple} ball \textbf{to the left of} the \textcolor{blue}{blue} ball?}}
    \label{fig:env}
\end{figure}

\section{Experiment}

In this section, we conduct experiments to evaluate the efficacy of the \methodname~method. The objective of the experiments is to address the following key questions:
(1) How does \methodname~enhance the acquisition of new skills in intelligent agents when compared to established baseline methods? (Refer to Section \ref{sec:exp_main_result}).
(2) How does \methodname~ground LLM in the environment? (Refer to Section \ref{sec:exp_analysis}).
(3) How are the rollouts generated by the fine-tuned LLM? (Refer to Section \ref{sec:exmp_rollouts}).
(4) What is the impact of the instruction-following fine-tuning of \methodname~on the overall performance of the algorithm? (Refer to Section \ref{sec:exp_ablation}).
We first introduce the environment used for experiments and the specific settings employed in our evaluation.

\subsection{Experimental Setting}
\label{sec:exp_setting}

\paragraph{Environment.} We conduct experiments within the CLEVR-Robot environment \cite{language_as_abstraction}, as depicted in Figure \ref{fig:env}. CLEVR-Robot, built on the MuJoCo physics engine \cite{mujoco}, facilitates object manipulation tasks. The environment involves an agent (silver point in the figure), and five manipulable balls. The state space is defined in $\mathbb{R}^{10}$, representing the positions of the balls, while the action space is $\mathbb{R}^{40}$, with each action is a one-hot vector dictating the movement of a specific ball in a given direction.

\paragraph{Novel task for evaluation.} To evaluate the efficacy of \methodname~and the learned policy's adaptability to new tasks, we define novel tasks and categorize them into three levels of complexity:

\begin{itemize}
    \item Rephrasing Goal: The agent performs same manipulation tasks but receives paraphrased natural language goals not present in the dataset. In rephrasing goal tasks, agent needs to complete the tasks existed in the offline dataset, while the natural language goals are different in expression. 
    \item Unseen (Easy): The agent is tasked with moving a specific ball to a given direction, e.g., \emph{Move the blue ball forward}. This level of tasks do not exist in the dataset, but require the LLM has a good understanding on the environmental data. To correctly generate the imaginary rollouts, the LLM must understand the meaningful of state, action, dynamics and their relation with the goals.
    \item Unseen (Hard): The agent faces tasks that are substantially different from those in the offline dataset, which require complex composition of behaviors, such as "Gather all balls together", "Move five balls to a straight line". For these tasks, LLM needs not only understand the meaning of environmental data, but also create novelty combination of each state to generate effective rollouts.
\end{itemize}

To utilize LLM to generate rollouts, we prompt ChatGPT \cite{chatgpt} to output the natural language goals that describe different levels of tasks, with examples provided in Appendix \ref{appsec:exm_novel_goal}. In our experiments, \methodname~trains policy on both offline dataset and the LLM-generated rollouts on these novel tasks, while baseline methods train the policy only on offline dataset.

\paragraph{Offline dataset.} The pre-collected offline dataset consists of 100,000 rollout-goal pairs, each corresponding to the task of rearranging a designated ball in a specified direction—forward, backward, left, or right—relative to a reference ball. For example, a natural language goal might be: "Move the green ball to the left of the blue ball." The dataset contains dense rewards indicating  The dataset encompasses $5 \times 4 \times 4 = 80$ unique configurations.
Following prior research \cite{talar}, we use 18 different natural language sentence patterns to describe each target configuration. 

When \methodname~fine-tunes LLM, we construct a training set comprising 400,000 trajectories, each rollout-goal pair in the offline dataset extending four trajectories: dynamics prediction, rollout explanation, rollout generation, and consequence prediction, as detailed in Section \ref{sec:method}. 
For training policies with offline RL, baseline methods only utilize the offline data (100,000 rollout-goal pairs) , while \methodname~generates additional 5600, 72400 and 1680 imaginary rollouts for rephrasing goal, unseen (easy) and unseen (hard) task, respectively. For each level of novel task, \methodname~trains the policy with both the offline dataset and the generated rollouts at this specific level, with same proportion of offline data and imaginary rollout in each training batch.

\paragraph{LLM and policy architecture.}
We utilize the Llama-2-7b-chat-hf model \cite{llama2} as the backbone LLM. The LLM undergoes training for 10 epochs with a batch size of 6, and we employ the fine-tuned model to generate rollouts corresponding to each task level. The offline RL training is replicated with three different random seeds to ensure robustness of the results. Details about hyper-parameters are provided in Appendix \ref{tab:hyper-parameters}.

\subsection{Main Results}

\label{sec:exp_main_result}

\textbf{Baselines.} We use several representative offline RL methods as comparison in our experiments. We briefly introduce them as follow: (1) Behavior Cloning (\textbf{BC}) adopts a supervised learning approach to replicate the actions found within the offline dataset. (2) Conservative Q-Learning (\textbf{CQL}) \cite{cql}, a prominent offline RL algorithm, constructs a cautious Q-function that ensures the policy's expected value does not exceed its true value.

\begin{figure}[h]
\centering
\subfigure[Task in Offline Data]{
\includegraphics[width=0.23\textwidth]{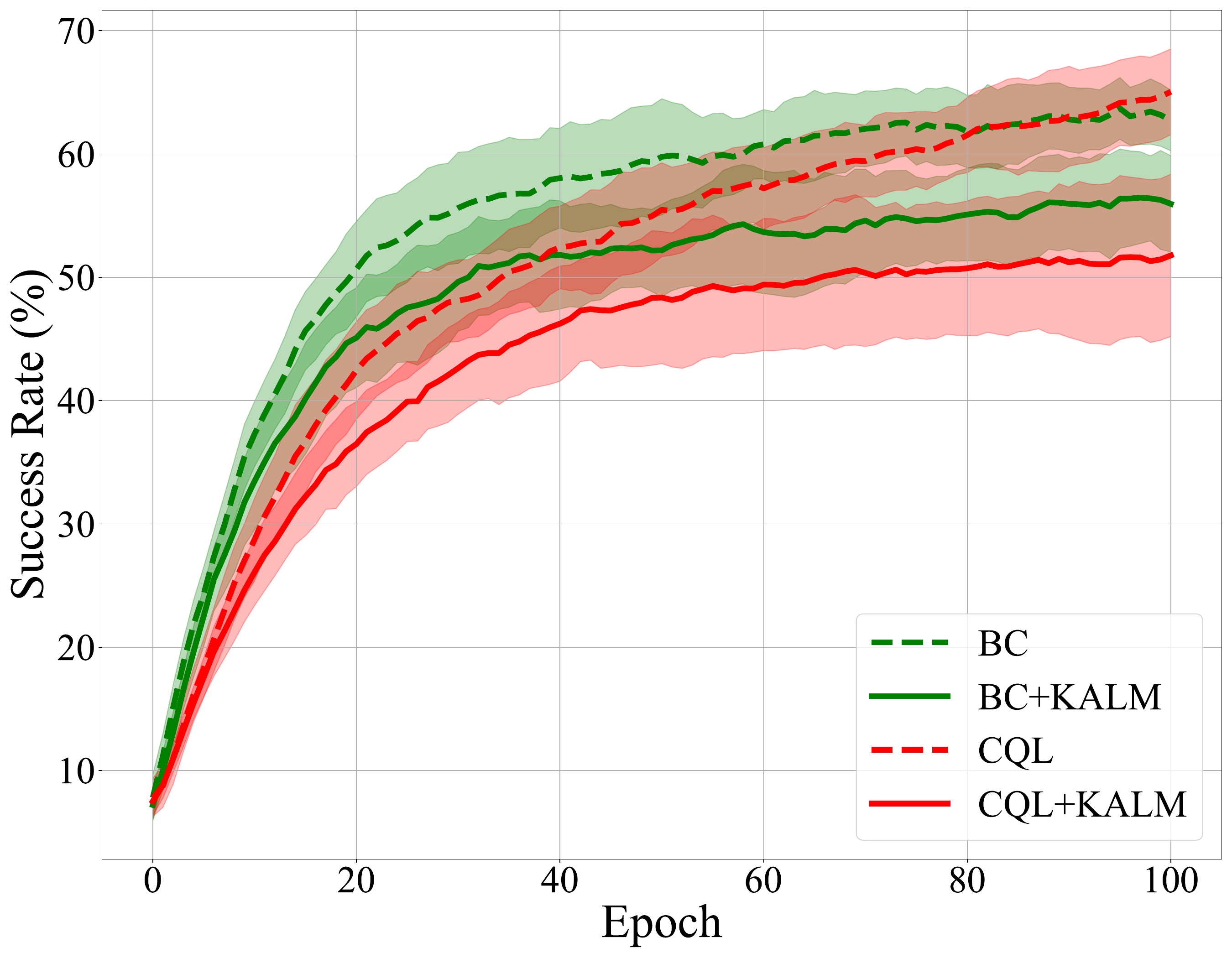}
\label{fig:task_in_od}
}
\subfigure[Rephrasing Goal]{
\includegraphics[width=0.23\textwidth]{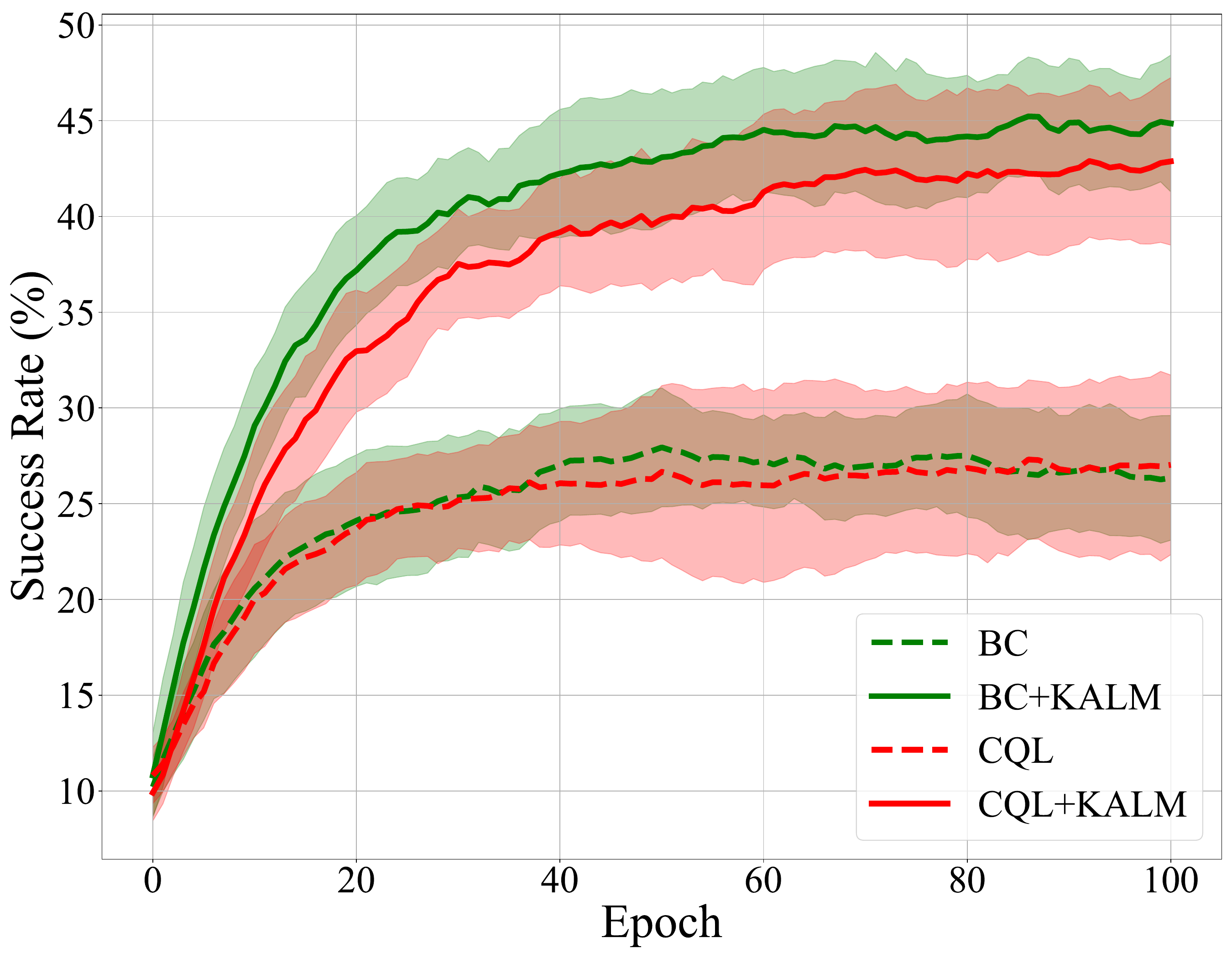}
}
\subfigure[Unseen (Easy)]{
\includegraphics[width=0.23\textwidth]{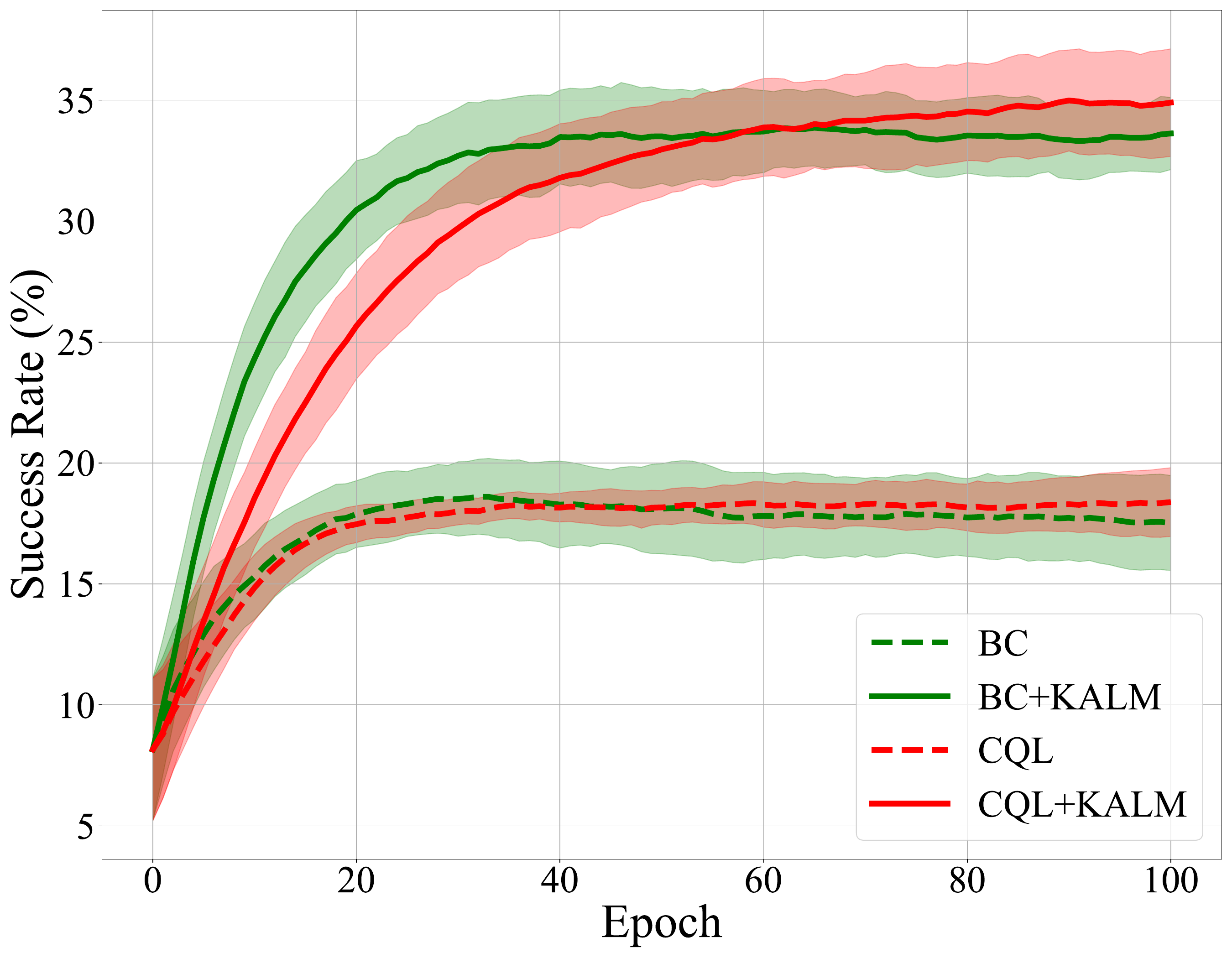}
}
\subfigure[Unseen (Hard)]{
\includegraphics[width=0.23\textwidth]{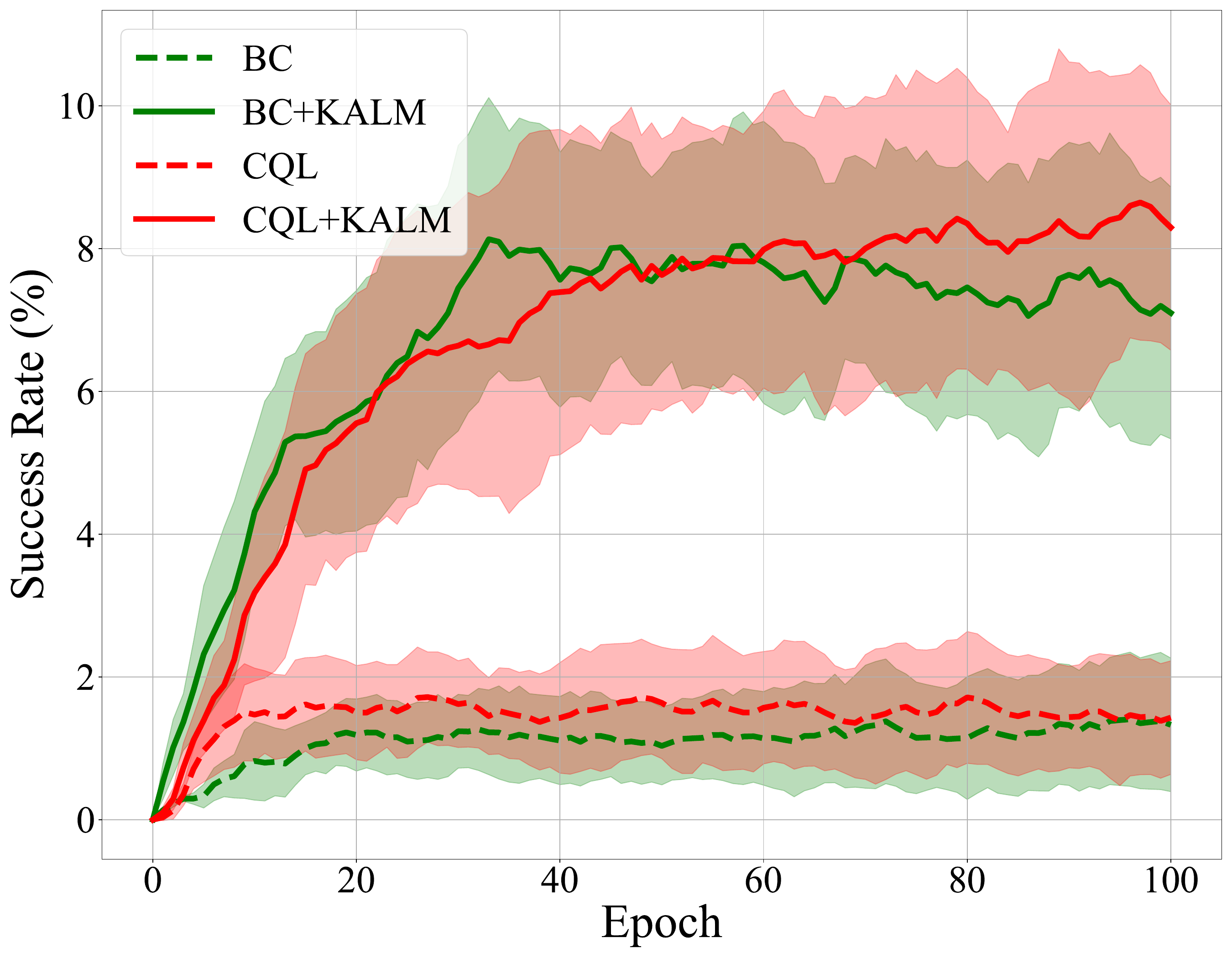}
}
\caption{Training curves of different methods on four types of tasks. The x-axis represents the number of training epochs, and the y-axis represents the success rate of completing the natural language goals for different types of tasks. The shaded area stands for the standard deviation over three random seeds.}
\label{fig:clevr_main_results}
\vspace{-2em}
\end{figure}

\textbf{Experimental results.} Figure \ref{fig:clevr_main_results} presents the comparative performance of various methods across different types of tasks. \methodname~trains the policy on the original offline data and the generated imaginary rollouts corresponding to corresponding level of the tasks. In Figure \ref{fig:task_in_od}, BC+\methodname~and CQL+\methodname~train the policy with offline data and generated rollouts for unseen (easy) tasks.
\methodname~consistently surpasses all baseline methods on all types of novel tasks. Notably, methods integrating offline data with generated rollouts demonstrate superior performance over those relying solely on offline data, underscoring the value of incorporating generated rollouts. On rephrasing goal tasks, CQL+\methodname~significantly outperforms both the BC and CQL methods, while achieving results on par with BC+\methodname. This indicates that methods trained exclusively on offline data exhibit limited generalizability to novel task expressions. In contrast, policies that incorporate generated rollouts exhibit marked performance enhancements.
Furthermore, we observe that CQL+\methodname and BC+\methodname~show enhanced performance on tasks existed in the offline dataset, indicating that the inclusion of generated rollouts not only preserves but potentially enhances performance on original tasks.
It is worth noting that \methodname~facilitates in unseen (hard) tasks over the baseline methods, reinforcing the utility of generated rollouts in acquiring novel skills.

\subsection{Performance of LLM Grounding}
\label{sec:exp_analysis}

Previous section demonstrates that \methodname~improves policy's performance on unseen novel tasks. In this section, we dive into the details of \methodname~running process, and examine why and how \methodname~improves the policy performance. Specifically, we demonstrate the quality of the generated rollouts and LLM's understanding on the original environment rollouts.

\begin{figure}[h]
\centering
\subfigure[Quality of the generated rollouts]{
\includegraphics[width = 0.59\linewidth]{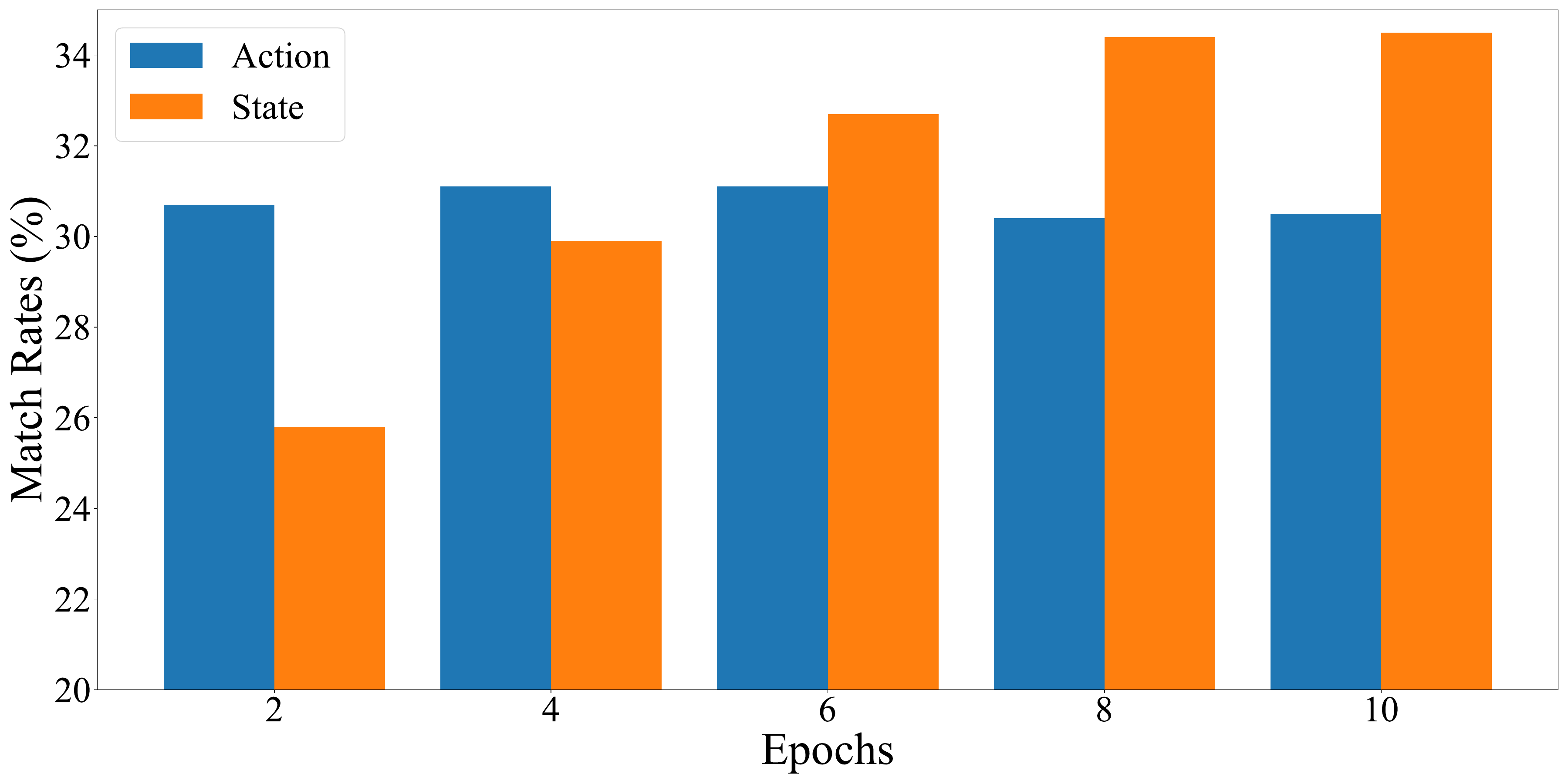}}
\subfigure[Accuracy of the rollout explanation]{
\includegraphics[width=0.38\textwidth]{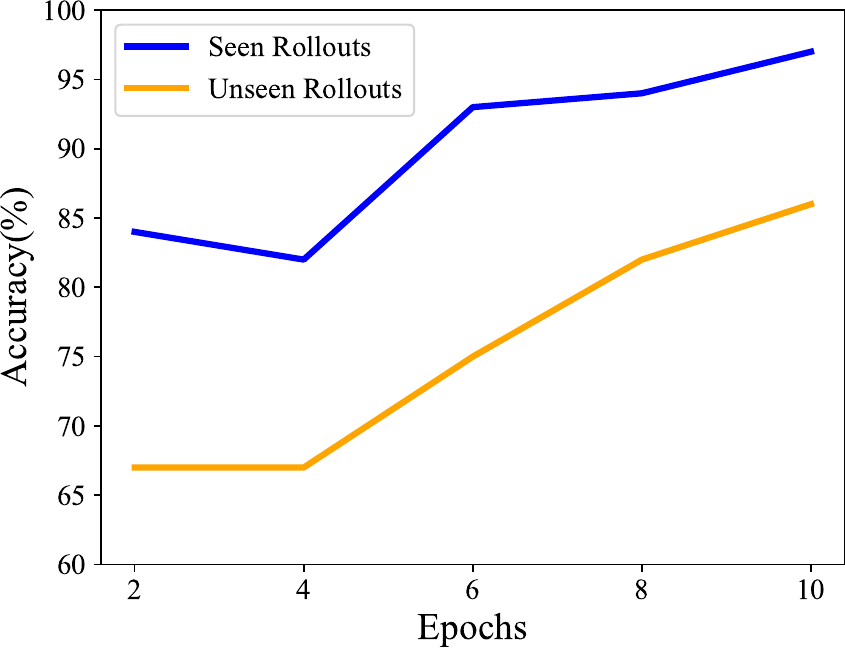}
\label{fig:acc_roll_explain}
}
\caption{We evaluate the efficacy of LLM grounding through two distinct metrics: \textbf{(a)} Quality of the generated rollouts, which is assessed by examining the alignment between the states and actions generated by the LLM and the labelled goals on unseen (easy) tasks. The horizontal axis represents the training epochs, while the vertical axis is the match rate of the generated states/actions. \textbf{(b)} Accuracy of the rollout explanation. To evaluate how does LLM understand the environmental rollouts, we present the LLM with both rollouts that LLM has seen or unseen during the training process. The LLM is then prompted to explain these rollouts. The accuracy is determined calculated by verifying whether it accurately identify and the objects and directions that align with the corresponding goal labels.}
\label{fig:generation_acc}
\end{figure}

\paragraph{Quality of the generated rollouts.} 
We examine the quality of the generated rollouts for unseen (easy) tasks. At this complexity, the agent's objective is to reposition a ball in a specified direction with a single action. The quality of the generated data is quantified by measuring the generation accuracy, which is determined by the match between the generated state/action and the corresponding natural language goal. For instance, given the goal "Move the red ball to the left," the accuracy is calculated by checking the alignment ratio of the generated state/action with this natural language goal. Figure \ref{fig:generation_acc} shows the results of the generation accuracy of five checkpoints during the training. The results show that while action generation accuracy remains constant at approximately 30\%, state generation accuracy improves as training progresses. These findings suggest that LLM works better in generating states than generating the actions. Besides, the LLM possesses a notable capacity to generate imaginary rollouts for novel tasks, despite not being explicitly trained on such tasks.

Understanding environmental dynamics is crucial for LLM to generate accurate predictions for novel tasks. We conduct experiments to assess the ability of the fine-tuned LLM to comprehend and explain environmental rollouts. The LLM was prompted to explain both seen and unseen rollout. As depicted in Figure \ref{fig:acc_roll_explain}, the LLM demonstrates a high level of explanatory accuracy, even after only two training epochs. These findings suggest that the LLM not only grasps the significance of numerical vector rollouts but also retains its original natural language processing proficiency after the fine-tuning.

\begin{figure}[h]
\centering
\includegraphics[width = 1.0\linewidth]{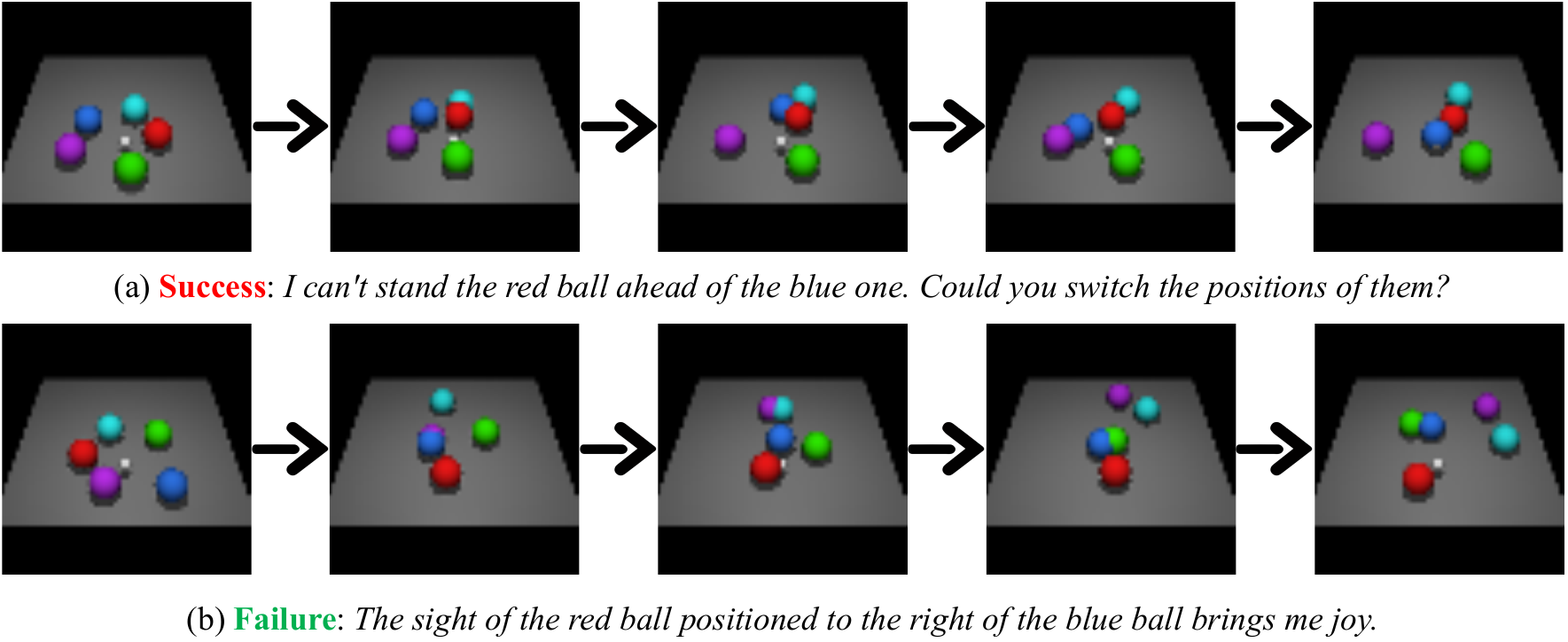}
\caption{Examples of the generated rollouts for rephrasing goal tasks (agent completes same tasks as offline data with rephrasing of natural language goals).}
\label{fig:GR_tau_level}
\end{figure}

\begin{figure}[h]
\centering
\includegraphics[width = 1.0\linewidth]{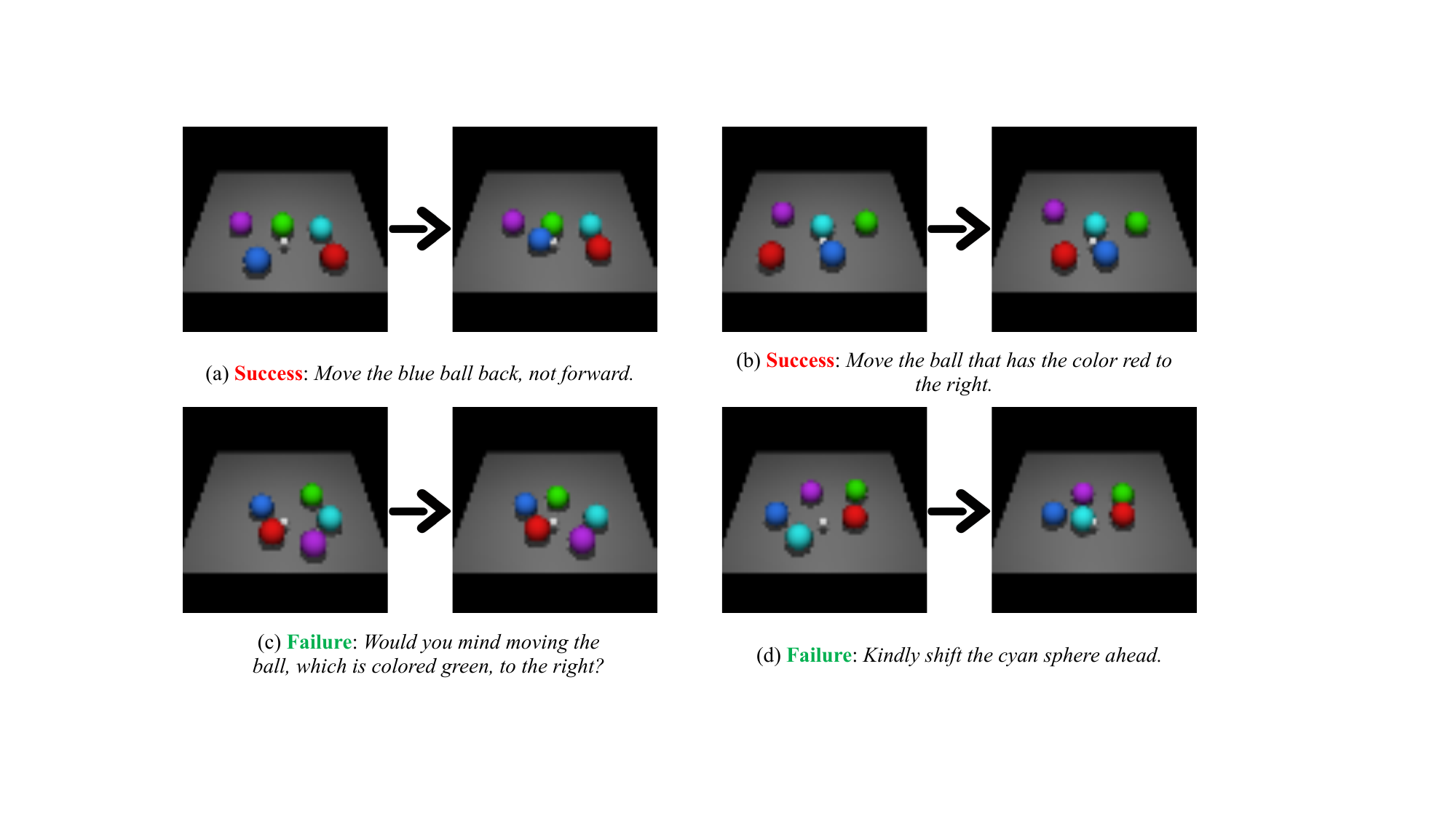}
\caption{Examples of the generated rollouts for unseen (easy) tasks (agent moves a specific ball to a given direction). Such tasks do not exist in the offline data.}
\label{fig:GR_step_level}
\end{figure}

\begin{figure}[h]
\centering
\includegraphics[width = 1.0\linewidth]{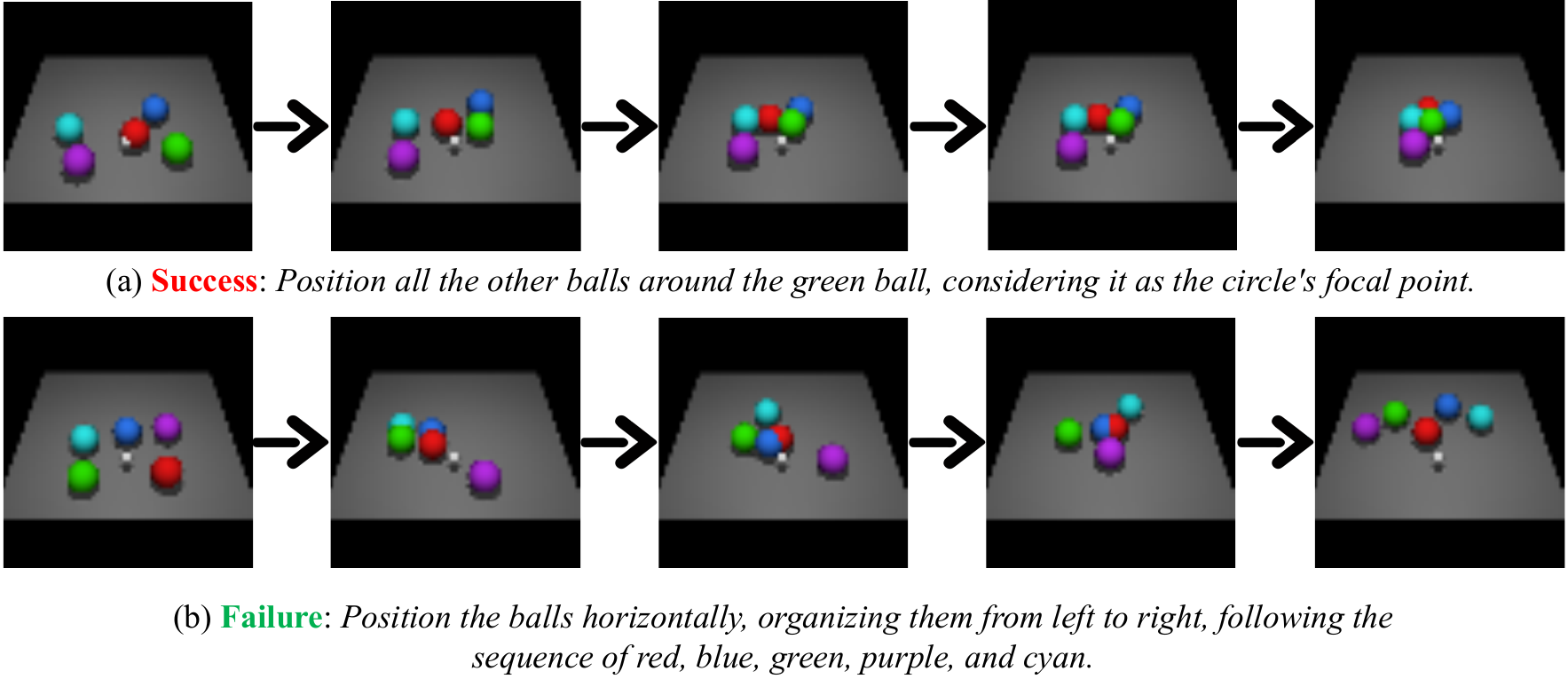}
\caption{Examples of the generated rollouts for unseen (hard) tasks (agent is tasked with completely new tasks).}
\label{fig:GR_task_level}
\end{figure}

\subsection{Examples of the Generated Rollouts}
\label{sec:exmp_rollouts}

We present the generated rollouts for various novel tasks in Figure \ref{fig:GR_tau_level}, \ref{fig:GR_step_level} and \ref{fig:GR_task_level}.
Overall, the generated rollouts conform to the basic laws of motion: LLM can correctly capture the objects, directions and the meaning of the novel goals. On rephrasing goal (Figure~\ref{fig:GR_tau_level}(a)): LLM successfully captures the semantics contained in the goals rather than directly perceiving the task goals through character matching. We also find that the performance of LLM might be sensitive to initial observation, as illustrated in Figure~\ref{fig:GR_tau_level}(b), the natural language goal is easy to understand while the LLM still fails due to the unfamiliar initial observation. On unseen (easy) tasks (Figure~\ref{fig:GR_step_level}(a)), the LLM adeptly captures the information from the goal, even the goal is substantially different from those encountered in the training set. However, from Figure~\ref{fig:GR_step_level}(b) we find that while the system demonstrates proficiency in color recognition, it encounters challenges in accurately discerning the trajectory of motion. Besides, we evaluate the performance of LLM from both the state and action perspectives due to the characteristic of this task (single-step decision-making). The evaluation results are shown in the Figure~\ref{fig:acc_roll_explain}, we observe that compared to modeling actions, LLM has a better ability to model states, which means LLM is more suitable for playing the role of environment rather than policy. On unseen (hard) (Figure~\ref{fig:GR_task_level}(a)), although the task requirements are more complex, requiring manipulations on multiple objects (whereas the training set of LLM only has examples about manipulations on single object), LLM is still able to complete the task. However, as illustrated in ~\ref{fig:GR_task_level}(b), as the difficulty of the task continues to increase, when the relative positional relationships of multiple objects need to be additionally considered, LLM begins to fail. Besides, we noticed that although LLM did not accurately arrange the items in a row according to the prescribed color order, there was a clear tendency to align the items in a row. We present more examples of the generated rollouts in Appendix \ref{appsec:add_examples}.

\subsection{Ablation Study}
\begin{figure}[h]
\centering
\includegraphics[width = 0.6\linewidth]{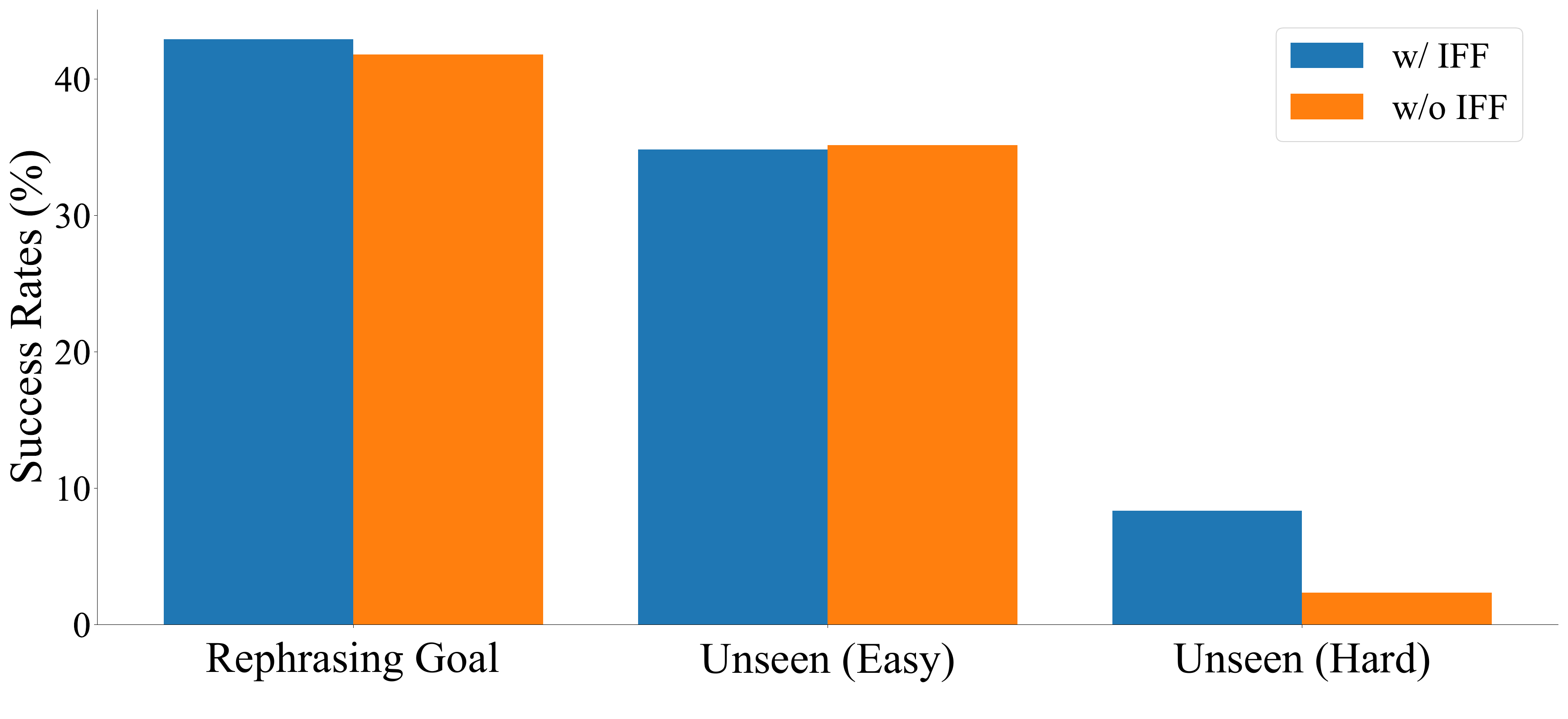}
\caption{Success rate of convergence of different LLM fine-tuning methods. The results are averaged on three different seeds and the last five epochs.}
\label{fig:ablation_grounding}
\end{figure}

We conduct ablation studies to assess the impact of instruction-following fine-tuning on LLM. For comparison purposes, we construct a dataset comprising sequences of the form $\{G,(s_0,a_0,\cdots)\}$ and fine-tune the LLM using a supervised approach without instruction-following prompts. As depicted in Figure \ref{fig:ablation_grounding}, the results indicate that LLMs fine-tuned with instruction-following prompts (denoted as 'w/ IFF') outperform those without such fine-tuning (denoted as 'w/o IFF'), particularly in generating superior rollouts that contribute to the development of high-quality policies. This enhancement is especially clear in tasks of unseen (hard).

\label{sec:exp_ablation}

\section{Conclusion \& Limitation}
This study investigates the adaptability of using a pre-trained LLM to directly generate environmental rollouts for novel tasks. We introduce a novel method, \methodname, which grounds LLM in the interactive environment and generates imaginary rollouts for novel skills. The offline RL techniques are then applied for skill acquisition for intelligent agents. The experiment results verify the effectiveness of \methodname~method, and the feasibility of utilizing LLM to generate imaginary rollouts for novel tasks. 
However, there are still some limitations. First topic is about the comprehensiveness of the experiments. For example, the LLM only processes vector state in this paper. To further evaluate the effectiveness of \methodname, future work can consider conducting experiments with tasks with visual observation, with an additional vision encoder to process visual data. Besides, the method is only evaluated on the CLEVR-Robot environment. It could be beneficial to conduct experiments on diverse tasks from different domain. 
Secondly, the LLM learns to generate both the state and the action corresponding to the novel tasks. This adds to the burden of the LLM to learn both dynamics and the behavior policy for executing certain skills.
Lastly, the study is limited to the language model domain, while there are many other multi-modal models, may further improve the applicability of \methodname~method.
We believe these interesting directions are worth further exploration for better leveraging the abundant knowledge implicitly embedded in the large models.

\clearpage

\bibliographystyle{unsrt}
\bibliography{references}

\clearpage

\appendix

\renewcommand{\thepart}{}
\renewcommand{\partname}{}

\part{\huge{\textbf{Appendix}}} %
\parttoc %
\clearpage

\section{Experiment Details}
\label{appsec:exp_details}

\subsection{Examples of Novel Goals}
\label{appsec:exm_novel_goal}

\subsubsection{Rephrasing Goal}
We use 40 different NL sentence patterns to express each goal configuration. For example, if we take a goal configuration such as "\textcolor{red}{red} ball and \textcolor{blue}{blue} ball", its corresponding NL instructions (i.e., eighteen NL sentence patterns) can be one of the following:
\begin{itemize}
    \item I can't stand the red ball ahead of the blue one. Could you switch the positions of them?
    \item The sight of the red ball ahead of the blue one bothers me. Can we reverse their order?
    \item I really dislike how the red ball is positioned in front of the blue ball. Could you exchange their places?
    \item It annoys me to see the red ball in front of the blue ball. Can we swap them around?
    \item Seeing the red ball ahead of the blue ball fills me with frustration. Let's switch them.
    \item The placement of the red ball in front of the blue ball is something I detest. Can you flip them?
    \item I can't bear to see the red ball positioned in front of the blue ball. Would you mind interchanging them?
    \item It irks me to have the red ball come in front of the blue ball. Could we trade their positions?
    \item The red ball being in front of the blue ball is something I can't tolerate. Let's switch them up.
    \item It really bothers me that the red ball precedes the blue ball. Can we swap their positions, please?
    \item Move the red ball gently to the left of the blue ball.
    \item Slowly nudge the red ball to the left side of the blue ball.
    \item Push the red ball towards the left of the blue ball at a leisurely pace.
    \item Slide the red ball to the left of the blue ball with a gentle touch.
    \item Gradually maneuver the red ball to the left side of the blue ball.
    \item Hasten the movement of the red ball to the left of the blue ball promptly.
    \item Expeditiously maneuver the red ball to the left side of the blue ball.
    \item Swiftly propel the red ball to the left, positioning it adjacent to the blue ball.
    \item Urgently shift the red ball to the left of the blue ball with rapidity.
    \item Accelerate the motion of the red ball towards the left of the blue ball expeditiously.
    \item Push the red sphere ahead of the blue sphere.
    \item Drive the red orb in front of the blue orb.
    \item Launch the red ball forward, preceding the blue one.
    \item Catapult the red sphere ahead of the blue sphere.
    \item Thrust the red sphere in front of the blue sphere.
    \item Propel the red-colored orb forward, leading the blue-colored orb.
    \item Send the red ball ahead of the blue ball.
    \item Fling the red sphere in front of the blue sphere.
    \item Hurl the red ball forward, preceding the blue ball.
    \item Cast the red sphere ahead of the blue sphere.
    \item The sight of the red ball positioned to the right of the blue ball brings me joy.
    \item It pleases me to observe the red ball situated on the right side of the blue ball.
    \item Seeing the red ball to the right of the blue ball fills me with happiness.
    \item I feel delighted witnessing the red ball located to the right of the blue ball.
    \item It brings me satisfaction to see the red ball positioned to the right of the blue ball.
    \item I am happy to notice the red ball situated on the right-hand side of the blue ball.
    \item Observing the red ball to the right of the blue ball brings me contentment.
    \item I am pleased by the arrangement of the red ball to the right of the blue ball.
    \item The red ball being on the right side of the blue ball gives me a sense of satisfaction.
    \item I find joy in the sight of the red ball being positioned to the right of the blue ball.
\end{itemize}

\subsubsection{Unseen (Easy)}
We use 80 different NL sentence patterns to express each goal configuration. For example, if we take a goal configuration such as "\textcolor{red}{red} ball", its corresponding NL instructions (i.e., eighteen NL sentence patterns) can be one of the following:
\begin{itemize}
    \item Move the ball backward, it's red.
    \item Push the red ball in reverse.
    \item Back up the red ball, please.
    \item Shift the red ball backwards.
    \item Can you move the red ball backwards?
    \item Retract the red ball, moving it backwards.
    \item Put the red ball in backward motion.
    \item Move the red ball back, not forward.
    \item Send the red ball backward, if you can.
    \item Maneuver the red ball to the rear.
    \item Drive the red ball backward, please.
    \item Pull the red ball back.
    \item Make the red ball move backwards.
    \item Shift the red ball rearward.
    \item Go backwards with the red ball.
    \item Execute a backward movement with the red ball.
    \item Make the red ball's position backward.
    \item Pull the red ball towards the back.
    \item Slide the red ball backwards.
    \item Propel the red ball backward, if possible.
    \item Kindly relocate the red sphere towards the left.
    \item Would you mind shifting the red orb to the left?
    \item I request that you move the red spherical object to the left.
    \item Could you please transfer the red ball towards the left?
    \item It would be appreciated if you could shift the red ball to the left.
    \item Please adjust the position of the red ball to the left.
    \item Kindly reposition the red ball to the left.
    \item I'd like you to move the red ball to the left, please.
    \item Please ensure the red ball is moved to the left.
    \item Could you relocate the red ball to the left?
    \item Please shift the red ball leftwards.
    \item Please slide the red ball over to the left.
    \item I need you to move the red ball leftward, please.
    \item Please nudge the red ball towards the left.
    \item Please push the red ball to the left.
    \item Would you kindly push the red ball towards the left?
    \item Kindly shift the red ball in the leftward direction.
    \item Could you move the red ball to the left side?
    \item It's required that you move the red ball towards the left.
    \item Please execute a leftward movement of the red ball.
    \item Kindly shift the red sphere ahead.
    \item Would you mind advancing the red orb?
    \item Can you push the red ball onward?
    \item Please nudge the red-colored sphere ahead.
    \item Kindly relocate the red-colored orb forward.
    \item Would you be so kind as to move the red-colored ball forward?
    \item Can you shift the red-colored sphere ahead?
    \item Please push the red-hued orb onward.
    \item Kindly advance the red-colored ball forward.
    \item Would you mind nudging the red sphere ahead?
    \item Can you move the red-colored ball forward?
    \item Please shift the red-toned orb onward.
    \item Kindly relocate the red-hued sphere forward.
    \item Would you be so kind as to push the red-colored ball forward?
    \item Can you nudge the red-colored sphere ahead?
    \item Please move the red-colored orb forward.
    \item Kindly advance the red-hued ball forward.
    \item Would you mind shifting the red sphere ahead?
    \item Can you push the red-colored orb onward?
    \item Please nudge the red-hued ball forward.
    \item Kindly relocate the red sphere to the starboard side.
    \item Move the red orb towards the right.
    \item Could you shift the red ball to the right?
    \item I request that you move the red ball to the right.
    \item Please shift the red ball to the right.
    \item Move the ball, which is red, to the right.
    \item Would you mind moving the ball, which is colored red, to the right?
    \item Kindly relocate the spherical object of red hue towards the right.
    \item Can you shift the ball, which happens to be red, to the right?
    \item I'd appreciate it if you could move the red ball to the right.
    \item Please adjust the position of the red ball to the right.
    \item Could you possibly move the red ball to the right?
    \item It would be great if you could move the red ball to the right.
    \item Kindly transfer the red-colored ball to the right.
    \item Move the ball that has the color red to the right.
    \item Would you kindly relocate the ball, specifically the red one, to the right?
    \item Please make the red ball move to the right.
    \item Can you shift the ball that's red to the right?
    \item Move the ball with the red hue to the right, please.
    \item Could you adjust the position of the ball, specifically the one that's red, to the right?
\end{itemize}

\subsubsection{Unseen (Hard)}
We designed 4 types of tasks for testing \methodname's performance on the completed unseen tasks: combination of two simple tasks, combination of 3 simple tasks, object arrangement task, and object collection task.
\begin{itemize}
    \item NL sentence patterns used in combination of simple tasks (Using "\textcolor{red}{red} ball \emph{behind} \textcolor{blue}{blue} ball" as goal configuration):
    \begin{enumerate}
        \item Push the red ball behind the blue ball.
        \item Move the red ball behind the blue ball.
        \item Keep the red ball behind the blue ball.
        \item Help me push the red ball behind the blue ball.
        \item Help me move the red ball behind the blue ball.
        \item Help me keep the red ball behind the blue ball.
    \end{enumerate}
    \item Combination of two simple tasks: Push the red ball behind the blue ball and move the green ball behind the purple ball.
    \item Combination of three simple tasks: Push the red ball behind the blue ball and move the green ball to the left of the purple ball and keep the cyan ball in front of the red ball.
    \item Object arrangement task
    \begin{enumerate}
        \item Place the balls horizontally, lining them up from left to right, in the order of red, blue, green, purple, and cyan.
        \item Set up the balls in a row from left to right, with red, blue, green, purple, and cyan in sequence.
        \item Arrange the balls in a line, moving from left to right, with red, blue, green, purple, and cyan.
        \item Position the balls horizontally, organizing them from left to right, following the sequence of red, blue, green, purple, and cyan.
        \item Line up the balls horizontally, sequencing them left to right as follows: red, blue, green, purple, and cyan.
        \item Order the balls in a row from left to right, with the sequence being red, blue, green, purple, and cyan.
        \item Arrange the balls in a horizontal line, starting from the left and proceeding to the right, with red, blue, green, purple, and cyan in order.
        \item Place the balls in a row horizontally, from left to right, in the sequence: red, blue, green, purple, and cyan.
        \item Set up the balls horizontally, arranging them in the order of red, blue, green, purple, and cyan from left to right.
        \item Line up the balls horizontally, sequencing them from left to right: red, blue, green, purple, and cyan.
        \item Position the balls in a horizontal row, ordering them left to right as follows: red, blue, green, purple, and cyan.
        \item Organize the balls horizontally, moving from left to right, with the sequence being red, blue, green, purple, and cyan.
        \item Place the balls in a line horizontally, arranging them from left to right, in the following order: red, blue, green, purple, and cyan.
        \item Set up the balls in a row horizontally, starting from the left and proceeding to the right, with red, blue, green, purple, and cyan in sequence.
        \item Arrange the balls in a horizontal line, sequencing them left to right as follows: red, blue, green, purple, and cyan.
        \item Order the balls in a horizontal row from left to right, with the sequence being red, blue, green, purple, and cyan.
        \item Position the balls horizontally, organizing them in the order of red, blue, green, purple, and cyan from left to right.
        \item Line up the balls horizontally, arranging them from left to right: red, blue, green, purple, and cyan.
        \item Place the balls in a horizontal row, ordering them left to right as follows: red, blue, green, purple, and cyan.
        \item Set up the balls in a row horizontally, moving from left to right, with the sequence being red, blue, green, purple, and cyan.
    \end{enumerate}
    \item Object collection task
    \begin{enumerate}
        \item Position all the other balls around the green ball, considering it as the circle's focal point.
        \item Use the green ball as the nucleus of the circle, arranging the rest around it.
        \item Let the green ball be the anchor of the circle, and arrange the others accordingly.
        \item Make the green ball the center of attention in the circle and rearrange the others accordingly.
        \item Arrange all other balls around the green one, treating it as the hub of the circle.
        \item Centralize the circle around the green ball, repositioning the others accordingly.
        \item Focus the circle around the green ball, adjusting the positions of the others.
        \item Orient the other balls around the green one, treating it as the central axis of the circle.
        \item Use the green ball as the reference point for the circle's arrangement, positioning the others around it.
        \item Position all the other balls around the green ball to create the circle.
        \item Arrange the other balls around the green ball, making it the center of the circle.
        \item Let the green ball dictate the layout of the circle, with the other balls positioned around it.
        \item Create the circle with the green ball as the center, arranging the others accordingly.
        \item Use the green ball as the pivot for the circle, arranging the other balls around it.
        \item Organize the circle around the green ball, adjusting the positions of the other balls.
        \item Centralize the arrangement of the circle around the green ball, repositioning the others.
        \item Treat the green ball as the central node of the circle and arrange the other balls accordingly.
        \item Position all the other balls around the green ball, with it as the focal point of the circle.
        \item Arrange the circle with the green ball at the center and the others positioned around it.
        \item Base the arrangement of the circle on the green ball, repositioning the others accordingly.
    \end{enumerate}
\end{itemize}

\subsection{Hyper-parameters in the Experiments}

\label{appsec:hyper-parameters}

The hyper-parameters for implementing \methodname~are presented in Table \ref{tab:hyper-parameters}. When implementing baseline methods, we use the same hyper-parameters, offline RL algorithms, and policy network architecture.

\begin{table}[h]
    \centering
    \caption{Hyper-parameters.}
    \begin{tabular}{c|c}
    \toprule
    \textbf{Hyper-parameters} & \textbf{Value}  \\   \toprule
    Discount Factor $\gamma$ & 0.99 \\ \midrule
    BC Batch Size & 100 \\ \midrule
    BC LR & 1e-3 \\ \midrule
    BC Imitation Weight & 0.5 \\ \midrule

    CQL Batch Size & 32 \\ \midrule
    CQL LR & 6.25e-5 \\ \midrule
    CQL Conservative Weight & 10.0 \\ \midrule

    MOBILE Batch Size & 1024 \\ \midrule
    MOBILE Actor LR & 1e-4 \\ \midrule
    MOBILE Critic LR & 3e-4 \\ \midrule
    MOBILE Dynamics LR & 1e-3 \\ \midrule
    MOBILE Number of Ensemble Q & 2 \\ \midrule
    MOBILE Target Q Penalty & 5.0 \\ \midrule
    MOBILE Dynamics Rollout Length & 1 \\ \midrule
    
    Feature Extrator Net & [$|\gS|$ + $|\mathcal{L}|$, 256, 256], ReLU \\
    \bottomrule
    \end{tabular}
    \label{tab:hyper-parameters}
\end{table}

\clearpage

\section{More Examples of Generated Rollouts}

\label{appsec:add_examples}

We present additional examples of the generated rollouts in Figure \ref{fig:appendix_GR_easy_level_succ}-\ref{fig:appendix_GR_hard_level_fail}.

\begin{itemize}
    \item Rephrasing Goal
    \begin{figure}[h]
        \centering
        \includegraphics[width = 1.0\linewidth]{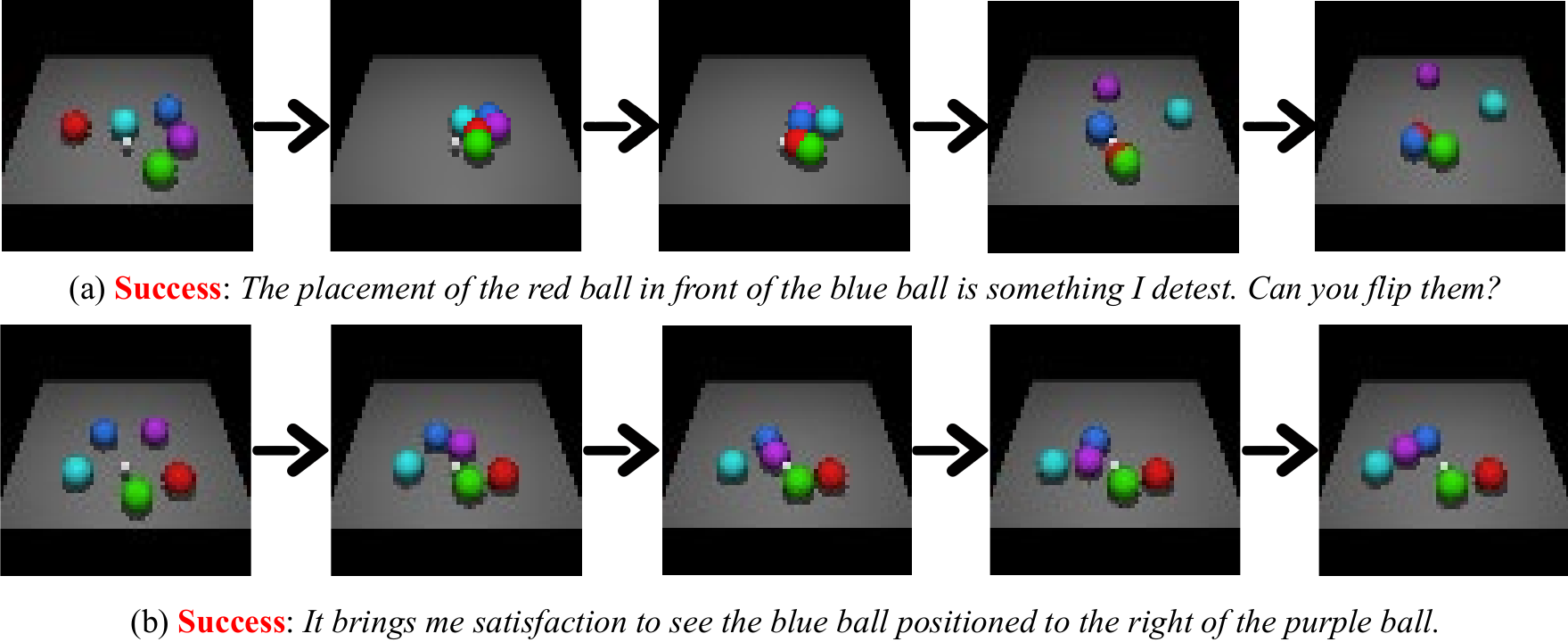}
        \caption{Success examples of the generated rollouts for rephrase goal tasks.}
        \label{fig:appendix_GR_easy_level_succ}
    \end{figure}
    \newpage
    \begin{figure}[h]
        \centering
        \includegraphics[width = 1.0\linewidth]{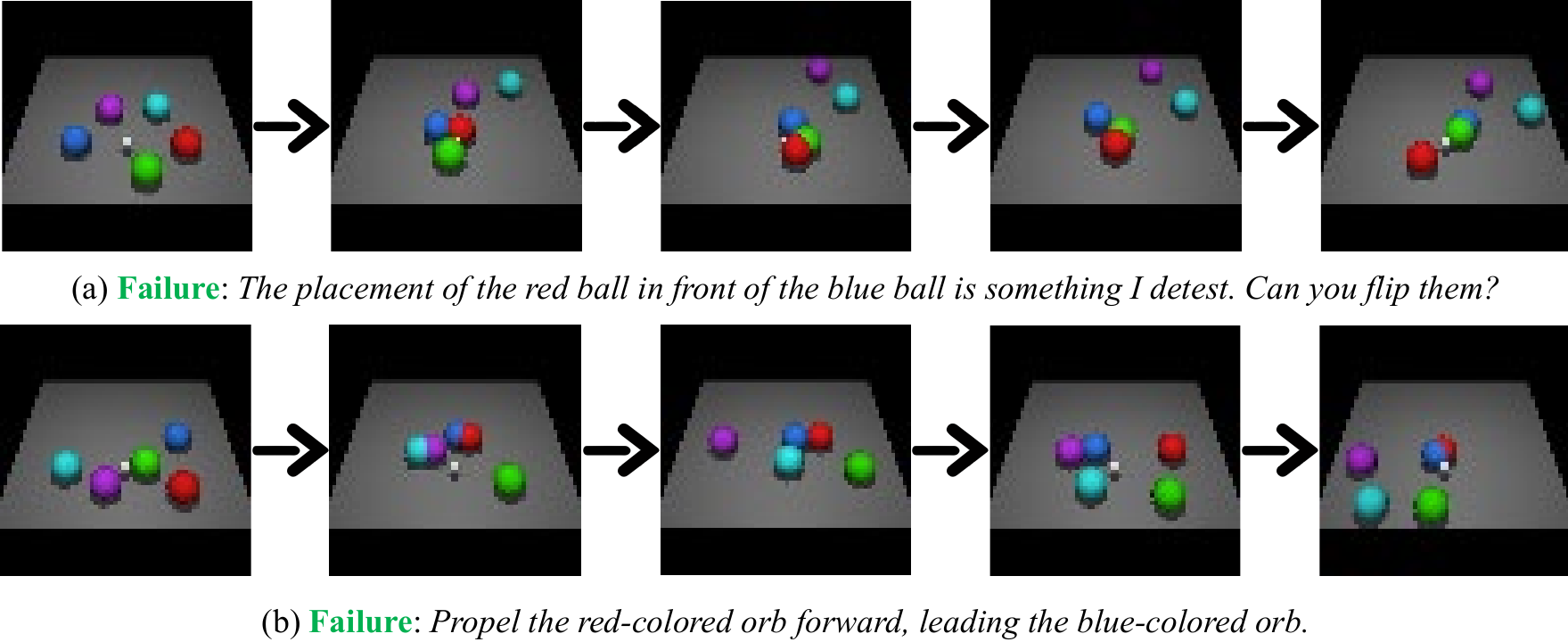}
        \caption{Extra failure examples of the generated rollouts for rephrase goal tasks.}
        \label{fig:appendix_GR_easy_level_fail}
    \end{figure}
    \item Unseen (Easy)
    \begin{figure}[h]
        \centering
        \includegraphics[width = 1.0\linewidth]{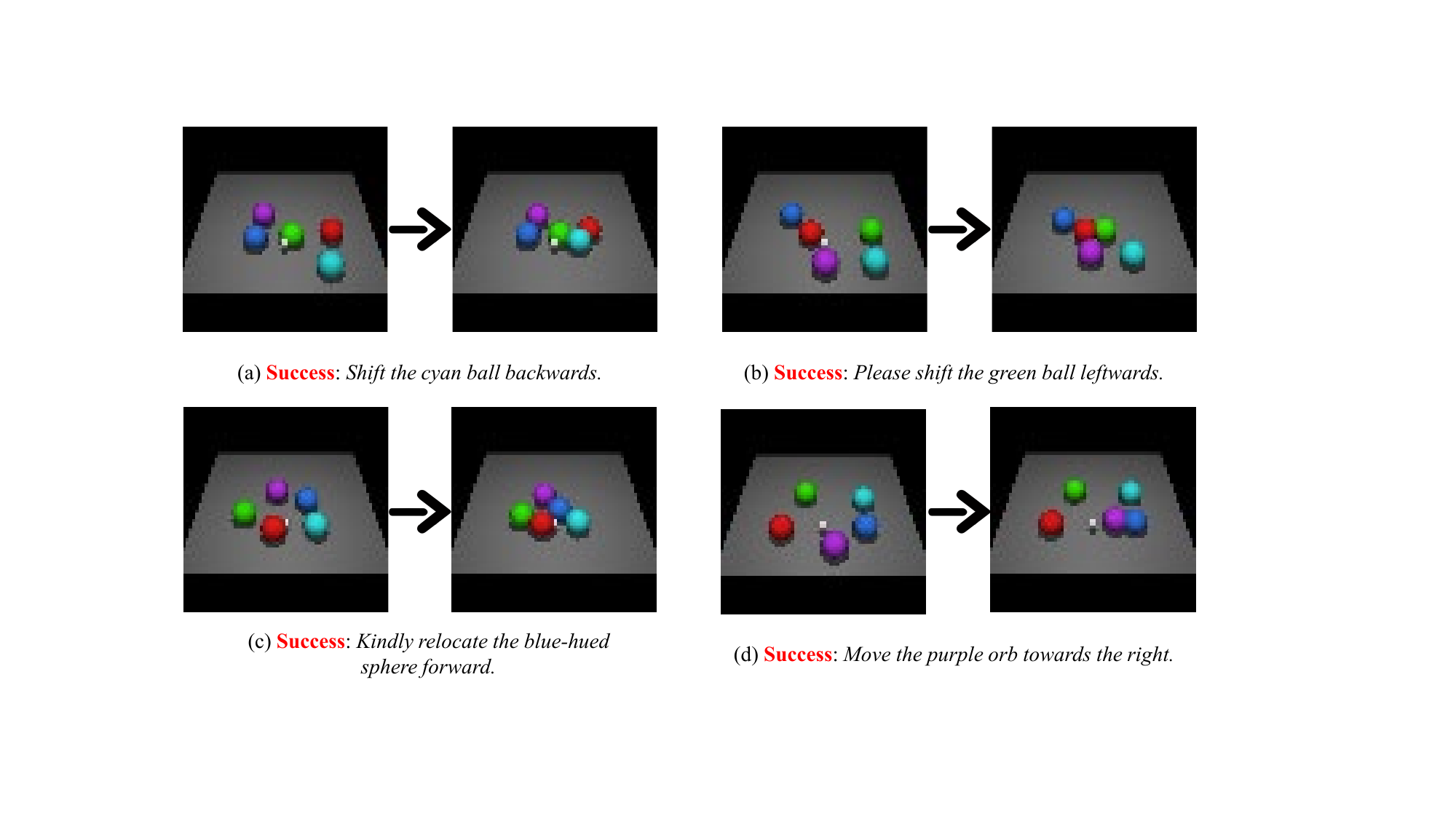}
        \caption{Extra success examples of the generated rollouts for unseen (easy) tasks.}
        \label{fig:appendix_GR_medium_level_succ}
    \end{figure}
    \newpage
    \begin{figure}[h]
        \centering
        \includegraphics[width = 1.0\linewidth]{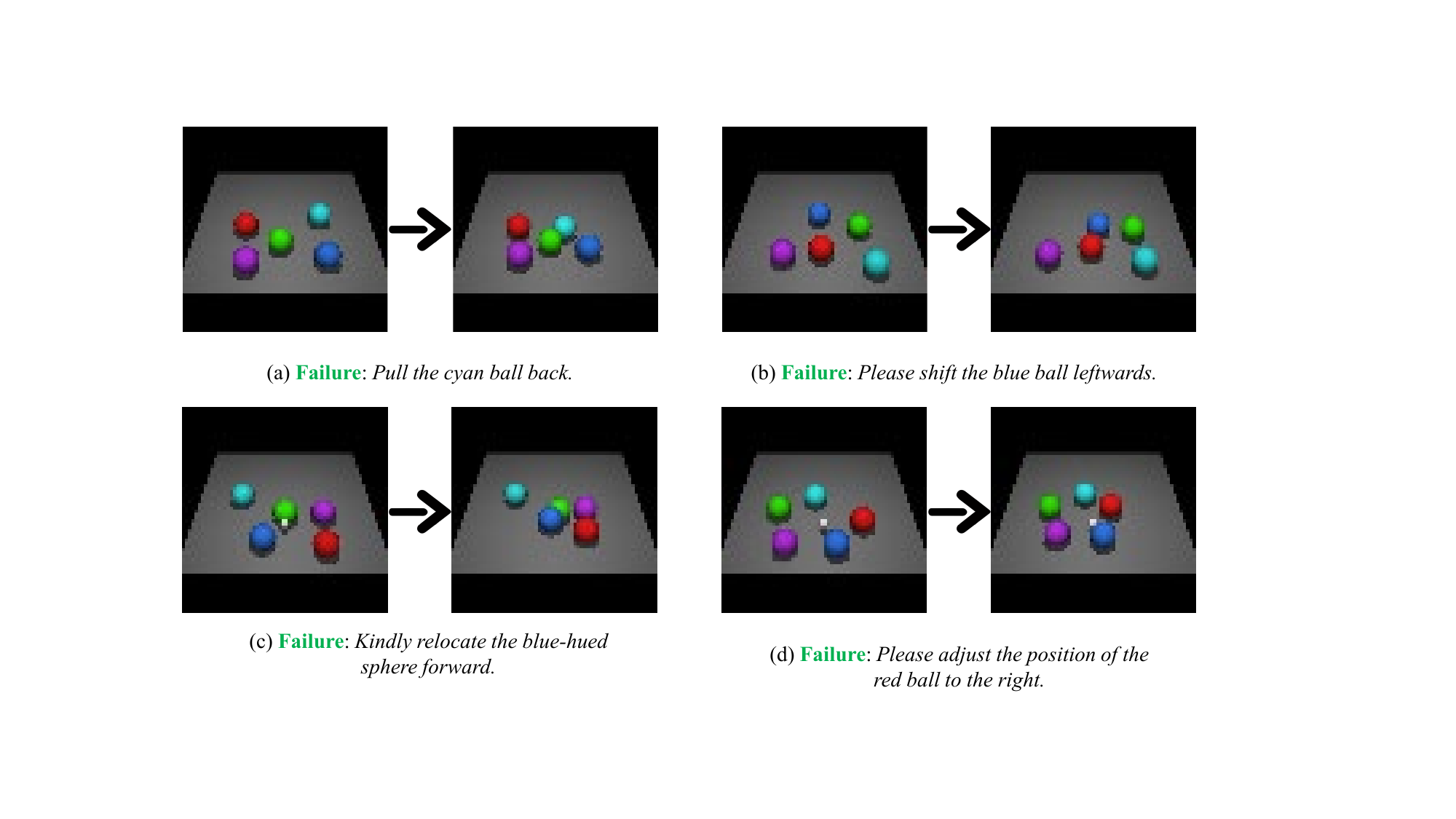}
        \caption{Extra failure examples of the generated rollouts for unseen (easy) tasks.}
        \label{fig:appendix_GR_medium_level_fail}
    \end{figure}
    \item Unseen (Hard)
    \begin{figure}[h]
        \centering
        \includegraphics[width = 1.0\linewidth]{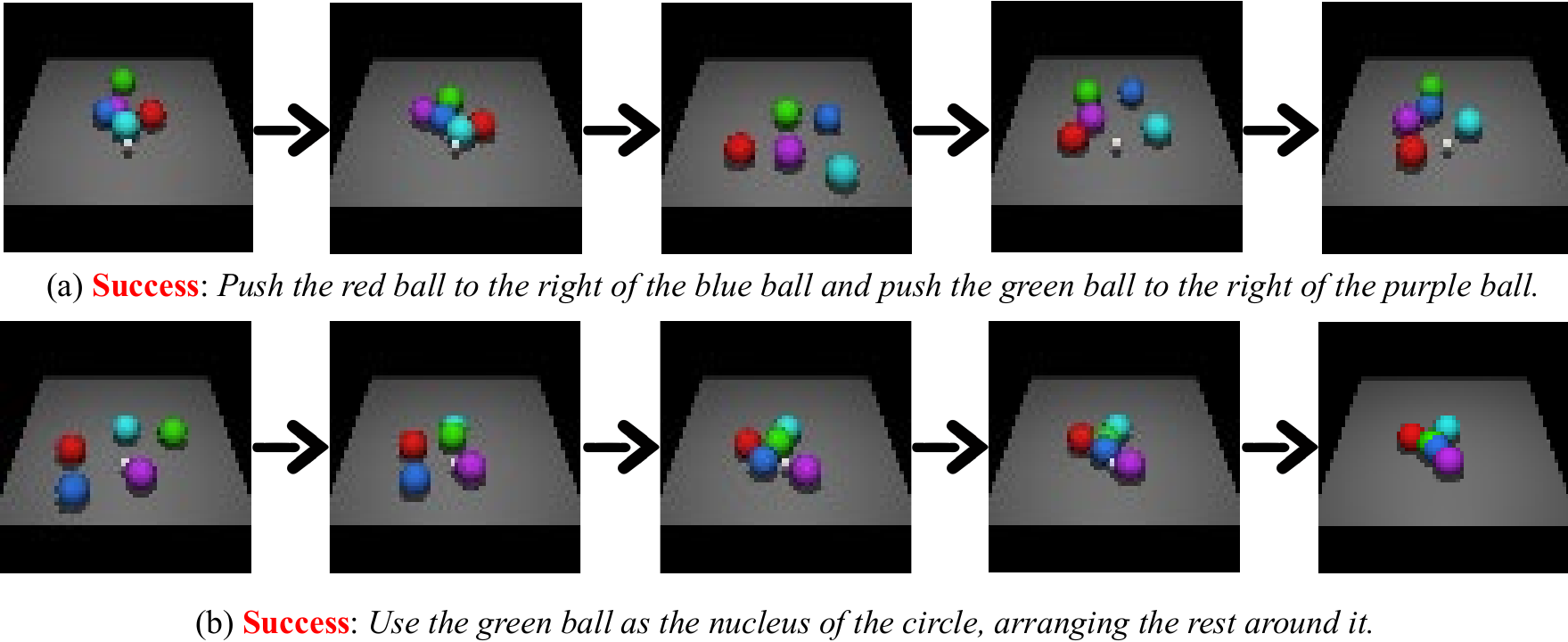}
        \caption{Extra success examples of the generated rollouts for unseen (hard) tasks.}
        \label{fig:appendix_GR_hard_level_succ}
    \end{figure}
    \newpage
    \begin{figure}[h]
        \centering
        \includegraphics[width = 1.0\linewidth]{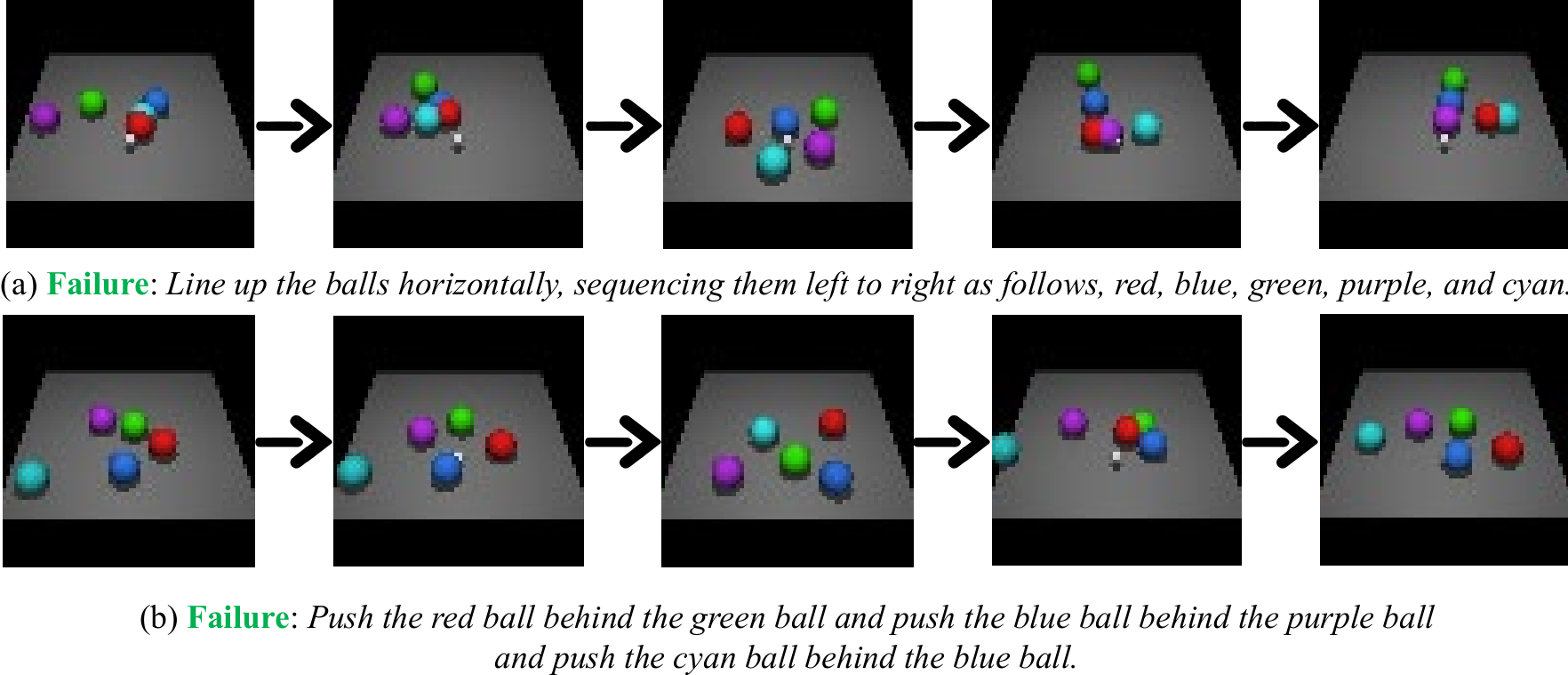}
        \caption{Extra failure examples of the generated rollouts for unseen (hard) tasks.}
        \label{fig:appendix_GR_hard_level_fail}
    \end{figure}
\end{itemize}

\section{Prompts for Instruction-following Fine-tuning}
\label{appsec:prompt}

\begin{itemize}
    \item Dynamics prediction: \emph{You are an expert in identifying dynamics change in the environment.
Current state is [$s_t$], after executing action [$a_t$], we get next state: [ANSWER].}
    \item Rollout to goal translation: \emph{Explain the following rollout for me. Rollout: [ROLLOUT]. Explanation: [ANSWER]}
    \item Goal to rollout translation: \emph{Please output the RL rollout corresponding to the following goal. Goal: [$G$]. Rollout: [ANSWER]}
    \item Consequence prediction: \emph{Initial state is [$s_0$], after completing the goal [$G$], we get terminal state [ANSWER].}
\end{itemize}

Here, [ANSWER] is the content that LLM should generate.

\end{document}